\documentclass[lettersize,journal]{IEEEtran}
\usepackage{amsmath,amsfonts}
\usepackage{array}
\usepackage[caption=false,font=normalsize,labelfont=sf,textfont=sf]{subfig}
\usepackage{textcomp}
\usepackage{stfloats}
\usepackage{url}
\usepackage{verbatim}
\usepackage{graphicx}
\usepackage{cite}
\usepackage{multirow} % 用于跨行
\usepackage{pifont} % 导入pifont宏包
\usepackage{colortbl} % 用于单元格颜色
\usepackage{xcolor}
\usepackage{algorithm}
\usepackage{algpseudocode}
\usepackage{caption}
\usepackage{booktabs}
\usepackage{booktabs}

\usepackage{xcolor}
\usepackage{graphicx}
\newcommand{\crbx}[2]{\scalebox{0.75}{\fcolorbox{#1}{#1}{\makebox[1.5ex][c]{\rule{0pt}{1.5ex}#2}}}}

\definecolor{tpami_blue}{RGB}{52,204,204}
\definecolor{tpami_gray}{RGB}{165,165,165}
\definecolor{tpami_red}{RGB}{192,0,0}
\definecolor{tpami_yellow}{RGB}{248,190,50}

\definecolor{crGeo}{RGB}{120,160,220}
\definecolor{crFunc}{RGB}{120,180,140}
\definecolor{crReg}{RGB}{220,150,120}

\newcommand{\Geo}{\crbx{crGeo}{\textcolor{white}{\textbf{\textsf{G}}}}}
\newcommand{\Func}{\crbx{crFunc}{\textcolor{white}{\textbf{\textsf{F}}}}}
\newcommand{\Reg}{\crbx{crReg}{\textcolor{white}{\textbf{\textsf{R}}}}}

\usepackage{array}
\newcolumntype{C}[1]{>{\centering\arraybackslash}p{#1}}
\newcolumntype{L}[1]{>{\raggedright\arraybackslash}p{#1}}

\usepackage{wasysym}   % For \CIRCLE
\usepackage{pdfpages}  % For PDF

\usepackage{hyperref}
\hypersetup{
	colorlinks=true,
	linkcolor=cyan,
	filecolor=blue,      
	urlcolor=black,
	citecolor=green,
}

\usepackage{etoolbox}
\patchcmd{\thebibliography}{\settowidth}
    {\setlength{\itemsep}{0pt plus 0.3ex}\settowidth}{}{}
\def\IEEEbibitemsep{0pt plus .5pt}
\let\oldthebibliography\thebibliography
\renewcommand{\thebibliography}[1]{%
  \oldthebibliography{#1}%
  \setlength{\itemsep}{\IEEEbibitemsep}%
  \scriptsize % 调整文献字体大小
}

\begin{document}

% Road Infrastructure Perception for Autonomous Driving: From Structured Sensing to Cooperative Extensions
% \title{Road Infrastructure Perception for Autonomous Driving: A Survey}
% \title{Road Infrastructure Perception in Autonomous Driving: Tasks, Representations, and Challenges}
\title{Road Infrastructure Perception for Autonomous Driving: Taxonomy, Representations, Tasks, and Challenges}

\author{
Guoyang~Zhao,
Weiqing~Qi,
Kai~Zhang,
Zhihai~Bi,
Tianshuai~Hu,
Zeying~Gong,
Dezong~Zhao,~\IEEEmembership{Senior~Member,~IEEE,}
and Jun~Ma,~\IEEEmembership{Senior~Member,~IEEE}
\IEEEcompsocitemizethanks{
\IEEEcompsocthanksitem Guoyang~Zhao, Weiqing~Qi, Kai~Zhang, Zhihai~Bi, and Zeying~Gong are with The Hong Kong University of Science and Technology (Guangzhou), Guangzhou 511453, China.
\IEEEcompsocthanksitem Tianshuai~Hu and Jun~Ma are with The Hong Kong University of Science and Technology, Hong Kong SAR, China.
\IEEEcompsocthanksitem Dezong Zhao is with the James Watt School of Engineering, University of
Glasgow, Glasgow G12 8QQ, U.K.
}
}



\IEEEaftertitletext{%
\vspace{-2.0em} % 缩小图与作者的距离
\noindent\begin{center}
\includegraphics[width=\linewidth]{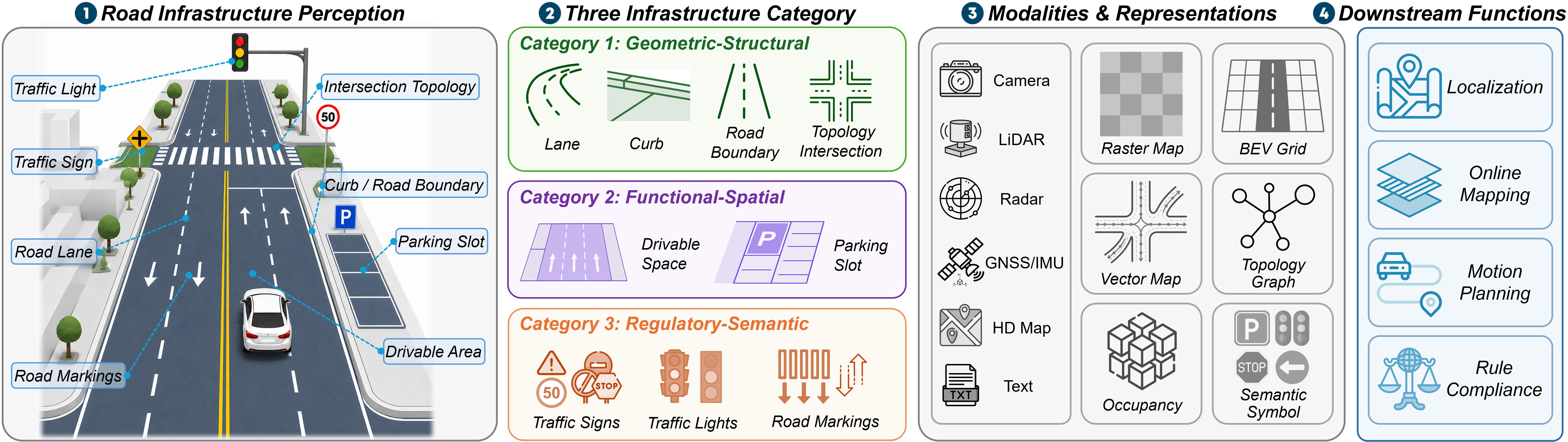}
\captionof{figure}{\textbf{Overview of road infrastructure perception for autonomous driving.}
This survey frames road infrastructure perception as a function-oriented process that converts sensing evidence and prior knowledge into structured road-scene representations. The proposed framework links geometric-structural, functional-spatial, and regulatory-semantic infrastructure to map-compatible perception and downstream decision-making.}
\label{cover-figure}
\end{center}
% \vspace{0.5em} % 视需要微调与摘要间距
\vspace{-5pt}
}

\maketitle

\begin{abstract}
Road infrastructure perception is a fundamental component of autonomous driving, as it recovers the structured road elements that define roadway geometry, delimit operational space, and encode traffic regulations. Although individual tasks such as lane detection, drivable-space estimation, traffic-sign recognition, and online map construction have been extensively studied, the field remains fragmented across task definitions, sensing assumptions, output representations, datasets, and evaluation protocols. This survey provides a function-oriented synthesis of road infrastructure perception. We first clarify its scope and organize existing studies into three complementary categories: geometric-structural, functional-spatial, and regulatory-semantic infrastructure. Based on this taxonomy, we review sensing modalities, scene representations, representative methods, benchmarks, and evaluation protocols, emphasizing the field’s evolution from image-space recognition toward metric, vectorized, topology-aware, occupancy-based, and rule-grounded scene modeling. We further examine how representation choices affect evaluation and downstream usability, revealing limitations of current benchmarks in relation reasoning, temporal consistency, robustness, and planning-oriented assessment. Finally, we discuss open challenges and future directions toward unified infrastructure representation, open-environment generalization, relation- and rule-aware reasoning, and downstream-oriented evaluation. By connecting tasks, representations, and evaluation under a common functional framework, this survey aims to establish road infrastructure perception as a coherent foundation for map-compatible and behavior-relevant autonomous-driving scene understanding.
\end{abstract}

% \begin{IEEEkeywords}
% Road infrastructure, scene understanding, perception, autonomous driving.
% \end{IEEEkeywords}


% \section{Introduction}
% \IEEEPARstart{T}{his} file is intended to serve as a ``sample article file''

% \title{Road Infrastructure Perception in Autonomous Driving: Tasks, Representations, and Challenges}

\section{Introduction}

Autonomous driving requires not only the perception of dynamic traffic participants, but also the structured understanding of road infrastructure that defines roadway geometry, constrains traversable space, and encodes traffic rules~\cite{badue2021self,hu2025vision}. While dynamic-object perception estimates the states and interactions of traffic participants, road infrastructure perception recovers the geometric, functional, and regulatory scaffold that governs autonomous driving behavior. Elements such as lane markings, road boundaries, curbs, drivable areas, parking slots, traffic signs, traffic lights, stop lines, crosswalks, and intersection layouts provide essential scene structure for lane-level localization, map alignment, motion planning, and rule-compliant decision-making~\cite{qi2024clrkdnet,paden2016survey,bresson2017simultaneous}. Therefore, infrastructure-oriented perception constitutes a fundamental component of autonomous driving systems.

Despite its importance, this field remains highly fragmented. Existing studies are typically organized around individual tasks, such as lane detection~\cite{zhang2024enhanced}, road-boundary and curb detection~\cite{hata2016feature}, drivable-area estimation~\cite{fritsch2013new}, parking slot detection~\cite{zhang2018vision}, traffic sign recognition~\cite{mogelmose2012vision}, traffic-light state understanding~\cite{Behrendt2017Deep}, road marking recognition~\cite{jayasinghe2022ceymo}, or HD map construction~\cite{elghazaly2023hdmaps}. Existing surveys have largely followed task-specific, modality-specific, or system-pipeline perspectives, leaving the cross-task commonalities of road infrastructure perception insufficiently synthesized~\cite{wen2022deep}. This fragmentation is not limited to task formulation, but also extends to sensing assumptions, coordinate systems, output representations, annotation protocols, evaluation metrics, benchmark datasets, and downstream interfaces. As a result, the field still lacks a coherent framework for organizing road infrastructure elements according to their driving functions, sensing requirements, representation forms, evaluation protocols, and roles in autonomous-driving systems~\cite{han2023collaborative}.

This gap is becoming increasingly consequential as autonomous driving perception shifts from object-centric recognition toward structured scene understanding. Bird's-eye-view (BEV) reasoning and online map construction make previously isolated infrastructure elements comparable within shared spatial representations~\cite{li2022hdmapnet}. Vectorized perception and topology-aware modeling further enable lanes, map elements, and traffic-related structures to be represented in compact, continuous, and relation-aware forms~\cite{wang2023openlanev2}. Meanwhile, HD maps are evolving from static priors toward online updating, infrastructure change detection, and long-term adaptation~\cite{elghazaly2023hdmaps}. Recent advances in foundation and vision-language models also motivate a shift from closed-set recognition toward semantic grounding and rule-aware interpretation of traffic signs, traffic lights, and road markings~\cite{radford2021learning}. 
These developments call for a dedicated survey that organizes road infrastructure perception not as a collection of isolated tasks, but as a unified research domain for behavior-relevant scene understanding.

In this survey, road infrastructure perception is defined as the detection, recognition, interpretation, and structured representation of road elements and regulatory states that organize roadway geometry, delimit functional driving space, or convey traffic rules. This definition places the scope of the field on driving function rather than physical appearance alone. As illustrated in Fig.~\ref{cover-figure}, road infrastructure perception is organized into three complementary categories: \textbf{\emph{geometric-structural infrastructure}}, which describes road layout and connectivity; \textbf{\emph{functional-spatial infrastructure}}, which determines operationally usable space; and \textbf{\emph{regulatory-semantic infrastructure}}, which encodes traffic rules and behavior constraints. Dynamic traffic participants are not treated as primary infrastructure elements unless they directly support the interpretation of road structure, accessible space, or regulatory semantics.

The contributions of this survey are fourfold. First, a unified framing of road infrastructure perception is established, clarifying its distinction from general scene understanding, object-centric perception, and isolated infrastructure-related tasks. Second, a function-oriented taxonomy is introduced to organize road infrastructure into geometric-structural, functional-spatial, and regulatory-semantic categories, providing a framework for comparing heterogeneous tasks. Third, representation design is highlighted as the key bridge from element-level perception to map-compatible and decision-relevant scene understanding. Fourth, dataset and evaluation gaps are analyzed to identify barriers between element-level benchmarks and autonomy-level infrastructure understanding, offering guidance for robust, relation-aware, map-compatible, and downstream-oriented perception systems.

\section{Definition, Scope, and Taxonomy}

\subsection{Definition and Scope}

In this survey, road infrastructure perception refers to the sensing, detection, recognition, and structured representation of road-scene elements that contribute to at least one of three driving-relevant functions: defining roadway geometry, constraining functional driving space, or conveying regulatory semantics~\cite{ulbrich2015defining,poggenhans2018lanelet2}. These elements are not grouped merely because they appear in road scenes, but because they provide essential information to autonomous driving: \textbf{\emph{geometric structure}}, which describes road layout, boundaries, and topology; \textbf{\emph{functional space}}, which determines traversable and operationally usable regions; and \textbf{\emph{regulatory semantics}}, which encode traffic rules and behavior-guiding instructions.

This functional definition distinguishes road infrastructure perception from general scene understanding. While general scene understanding may cover all visible entities, road infrastructure perception focuses on elements that organize roadway structure, delimit vehicle-accessible space, or regulate vehicle behavior. Its scope is therefore determined by driving function rather than by physical appearance alone, making it directly relevant to localization, map reasoning, motion planning, and rule-compliant autonomy~\cite{poggenhans2018lanelet2, ning2024novel}.

Accordingly, the primary scope of this survey includes lane markings, lane boundaries, road boundaries, curbs, medians, shoulders, lane connectors, intersection layouts, drivable areas, free space, non-drivable regions, parking slots, traffic signs, traffic lights, stop lines, crosswalk markings, directional arrows, road text, lane-use control markings, and other rule-bearing road markings~\cite{zhang2022deep,zhang2018road,qian2020dlt,tabernik2019deep}. Dynamic traffic participants, such as vehicles, pedestrians, and cyclists, are outside the main scope unless they directly support infrastructure understanding. General background objects, including buildings, trees, and roadside clutter, are also excluded except when they provide auxiliary context for infrastructure perception.

The boundary of this survey is therefore not determined by a simple static--dynamic distinction. Semi-static or temporary infrastructure-like elements, such as traffic cones, temporary barriers, construction signs, lane-closure boards, variable message signs, and road studs, are considered only when they explicitly modify road organization, accessible space, or regulatory constraints~\cite{ulbrich2015defining}. Otherwise, they fall under general object or obstacle perception. In this sense, an element belongs to road infrastructure perception if it contributes to the geometric organization, functional accessibility, or regulatory semantics of the roadway.

\begin{figure}[t]
    \centering
    \includegraphics[width=0.5\textwidth]{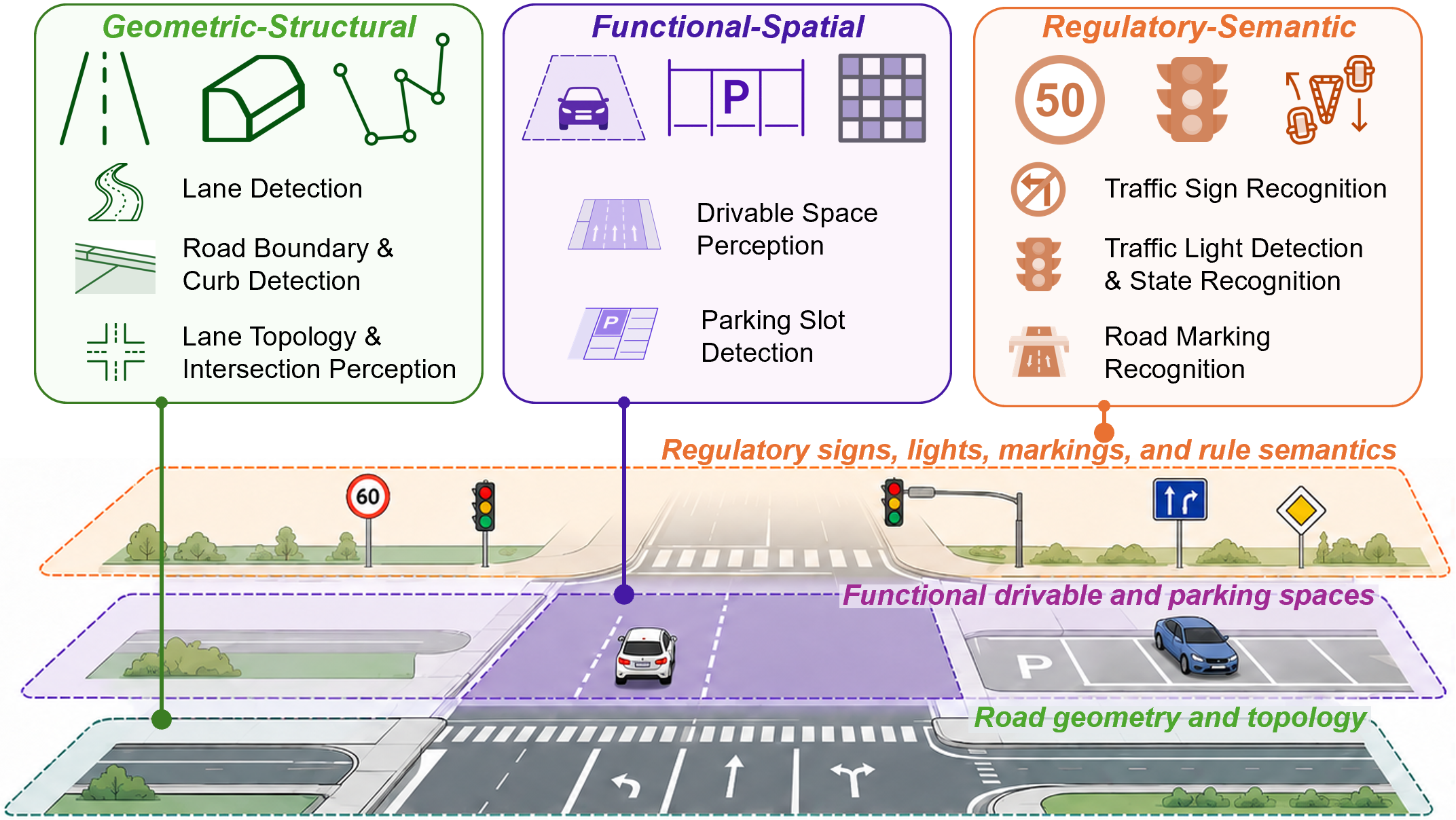}
    \vspace{-12pt}
    \caption{\textbf{Visual taxonomy of road infrastructure perception.}
The proposed taxonomy organizes road infrastructure perception into three complementary categories according to the dominant driving-relevant information recovered from road scenes: geometric-structural, functional-spatial, and regulatory-semantic infrastructure.}
    \label{fig:taxonomy}
    \vspace{-15pt}
\end{figure}

\subsection{Taxonomy of Road Infrastructure Perception}

To organize this field systematically, we propose the taxonomy shown in Fig.~\ref{fig:taxonomy}, which categorizes road infrastructure perception according to the dominant type of driving-relevant information recovered from road scenes, rather than by sensor configuration, object appearance, or annotation format~\cite{poggenhans2018lanelet2,qian2020dlt}. At the category level, road infrastructure is divided into geometric-structural infrastructure, functional-spatial infrastructure, and regulatory-semantic infrastructure. At the task level, each category is associated with representative perception problems and infrastructure elements, as summarized in Table~\ref{tab:infra_taxonomy}. These categories are not strictly exclusive, since many infrastructure elements simultaneously carry geometric, functional, and semantic implications. Instead, they provide an analytical framework for comparing heterogeneous tasks under a common infrastructure-oriented perspective.

\begin{table*}[t!]
\renewcommand\arraystretch{1.18}
\caption{\textbf{Standardized taxonomy of road infrastructure perception tasks for autonomous driving.}}
\centering
\scriptsize
\setlength{\tabcolsep}{1.2mm}
% \begin{tabular}{p{0.07\textwidth} p{0.24\textwidth} p{0.34\textwidth} p{0.31\textwidth}}
\resizebox{\textwidth}{!}{
\begin{tabular}{p{1.2cm} p{4.6cm} p{5.9cm} p{5.7cm}}
\hline
\rowcolor{gray!10}
\textbf{Category} 
& \textbf{Standardized Task} 
& \textbf{Representative Elements} 
& \textbf{Downstream Relevance} \\
\hline

\multirow{3}{=}{Geometric-Structural}
& (1) Lane Detection
& Lane markings, boundaries, dividers, connectors
& Lane keeping, localization, planning \\

& (2) Road Boundary and Curb Detection
& Road boundaries, curbs, edges, medians, shoulders
& Road-edge constraint, map alignment, safety \\

& (3) Lane Topology and Intersection Perception
& Lane connectivity, merge/diverge, intersection layout
& Routing, maneuver reasoning, online mapping \\
\hline

\multirow{2}{=}{Functional-Spatial}
& (4) Drivable Space Perception
& Drivable, free, non-drivable, accessible regions
& Local planning, collision avoidance, traversability \\

& (5) Parking Slot Detection
& Parking slots, boundaries, structured spaces
& Automated parking, low-speed maneuvering, local maps \\
\hline

\multirow{3}{=}{Regulatory-Semantic}
& (6) Traffic Sign Recognition
& Warning, prohibition, mandatory, speed-limit signs
& Rule compliance, speed control, route constraints \\

& (7) Traffic Light and State Recognition
& Traffic lights, signal states, lane-relevant signals
& Right-of-way, intersection decisions \\

& (8) Road Marking Recognition
& Stop lines, crosswalks, arrows, road text, lane-use markings
& Maneuver constraints, stopping, lane-use reasoning \\
\hline
\end{tabular}}
\vspace{-12pt}
\label{tab:infra_taxonomy}
\end{table*}

\textbf{\emph{Geometric-structural infrastructure}} comprises elements that define the physical layout, spatial boundaries, and topological organization of the road~\cite{zhang2022deep,zhang2018road}. This category includes lane markings, lane boundaries, road boundaries, curbs, medians, shoulders, lane connectors, and intersection layouts. In task form, it is instantiated through \emph{Lane Detection}, \emph{Road Boundary and Curb Detection}, and \emph{Lane Topology and Intersection Perception}~\cite{poggenhans2018lanelet2}. These tasks require representations that preserve geometric continuity, boundary consistency, metric accuracy, and topological validity.

\textbf{\emph{Functional-spatial infrastructure}} refers to elements that determine where a vehicle can, should, or must not move~\cite{qian2020dlt}. Representative elements include drivable areas, non-drivable regions, accessible driving space, parking slots, and parking boundaries. Unlike geometric-structural infrastructure, which primarily describes road form, functional-spatial infrastructure describes the safe operational availability of space. This category is represented by \emph{Drivable Space Perception} and \emph{Parking Slot Detection}, which connect infrastructure perception with traversability reasoning, low-speed maneuvering, and planning-compatible spatial representation~\cite{suhr2021end}.

\textbf{\emph{Regulatory-semantic infrastructure}} includes elements that convey traffic rules, legal constraints, or behavior-guiding instructions~\cite{tabernik2019deep,possatti2019traffic}. Representative elements include traffic signs, traffic lights, stop lines, crosswalk markings, directional arrows, road text, lane-use control markings, and other rule-bearing markings~\cite{morrissett2020review}. This category is instantiated through \emph{Traffic Sign Recognition}, \emph{Traffic Light Detection and State Recognition}, and \emph{Road Marking Recognition}. Compared with the previous two categories, it depends more heavily on high-resolution visual perception, symbolic interpretation, lane-level association, temporal state understanding, and semantic or rule-aware reasoning.

\subsection{Taxonomy as an Organizing Framework}

The proposed taxonomy serves as the conceptual backbone of this survey. It is not merely a descriptive list of road objects, but a common coordinate system for analyzing infrastructure perception across sensing modalities, scene representations, task formulations, datasets, evaluation protocols, and downstream autonomy requirements. Geometric-structural infrastructure emphasizes continuity, boundary localization, topology, and map alignment; functional-spatial infrastructure emphasizes traversability, accessibility, and planning-compatible spatial inference; and regulatory-semantic infrastructure emphasizes semantic resolution, symbolic grounding, lane relevance, and rule interpretation.

This taxonomy also clarifies why existing studies, although often developed under separate task names, share deeper methodological connections. At the sensing level, different categories rely on different forms of infrastructure evidence, such as visual semantics, metric geometry, spatial continuity, and map alignment. At the representation level, they explain the evolution from dense masks toward BEV grids, vectorized primitives, topology graphs, occupancy fields, and semantic-symbolic attributes. At the benchmark level, they reveal why existing datasets and metrics remain fragmented and why representation-aware and downstream-oriented evaluation is necessary. In the following sections, we use this taxonomy to organize the discussion of sensing modalities, scene representations, core tasks, benchmarks, challenges, and future research directions.

\section{Sensing Modalities and Scene Representations}

\begin{figure}[t]
    \centering
    \includegraphics[width=0.5\textwidth]{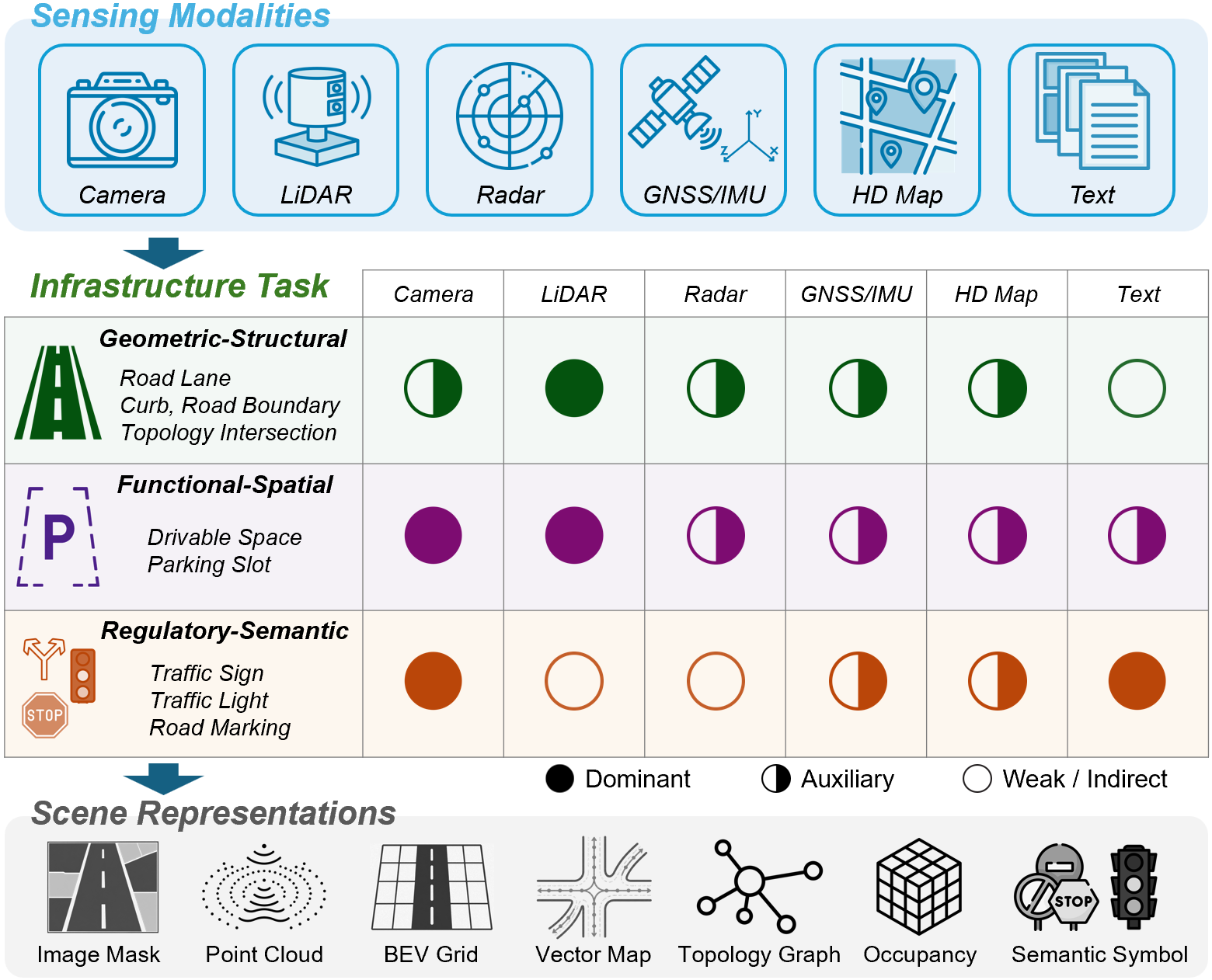}
    \vspace{-12pt}
    \caption{\textbf{Relationship between sensing modalities and scene representations in road infrastructure perception.} Different infrastructure categories impose distinct sensing requirements and representation preferences. Vision is dominant for texture- and semantics-rich elements, LiDAR is critical for geometry- and boundary-oriented perception, while GNSS/IMU and map priors mainly support alignment, temporal consistency, and structured reasoning.}
    \label{fig:sensing_representation}
    \vspace{-15pt}
\end{figure}

\subsection{Infrastructure-Oriented Sensing Evidence}

Road infrastructure perception can be understood as a process that transforms heterogeneous sensing evidence into structured scene representations for autonomous driving~\cite{cress2023intelligent,zhao2025fisheyedepth}. Unlike object-centric perception, which mainly emphasizes object appearance, category, and motion, infrastructure perception must recover roadway geometry, functional driving space, and regulation-bearing semantics in forms that can support localization, mapping, planning, and rule-compliant decision-making~\cite{chen2024end,dal2025joint}. Therefore, sensing modalities should be compared not only as hardware inputs, but also as complementary sources of infrastructure evidence~\cite{hakim2024stc,shi2024grid}. Different infrastructure categories impose distinct sensing requirements and representation preferences. Fig.~\ref{fig:sensing_representation} provides a visual summary of this relationship between sensing modalities, infrastructure categories, and scene representations.

For \textbf{\emph{geometric-structural infrastructure}}, including lanes, road boundaries, curbs, lane topology, and intersections, the key sensing requirement is to recover spatial continuity, boundary location, elevation discontinuity, and topological organization~\cite{ma2025monocular}. Cameras are effective when lane markings or road boundaries exhibit strong visual contrast, whereas LiDAR provides more direct metric evidence for curbs, road edges, height changes, weakly marked boundaries, and structural road layout~\cite{arnold2019survey, zhao2024learning}. GNSS/IMU and map priors further contribute to ego-motion estimation, temporal stabilization, and map alignment, which are particularly important for topology-aware and map-coupled perception~\cite{kuutti2018survey}.

For \textbf{\emph{functional-spatial infrastructure}}, such as drivable space and parking slots, sensing is less about recognizing isolated objects than inferring the operational availability of space. Visual sensors provide road texture, scene context, and marking-level cues, while LiDAR and depth sensors contribute geometric evidence for separating traversable surfaces from sidewalks, curbs, barriers, and other non-drivable structures~\cite{muhammad2022vision,liu2024omnicolor}. In this category, sensing is closely coupled with planning-oriented spatial inference, because the output must indicate where the vehicle can safely move, stop, or maneuver~\cite{xie2023circular}.

For \textbf{\emph{regulatory-semantic infrastructure}}, including traffic signs, traffic lights, and rule-bearing road markings, high semantic resolution is essential. Cameras remain the dominant modality because these elements are often small, visually fine-grained, color-dependent, and symbolically meaningful~\cite{yang2024traffic}. LiDAR and radar usually provide auxiliary geometric or robustness cues rather than direct semantic recognition. Map priors, lane-level context, and language-grounded perception may further support semantic disambiguation, lane relevance estimation, and rule interpretation across different regions~\cite{guo2024signparser}.

Overall, different sensing modalities provide complementary infrastructure evidence rather than interchangeable observations. Vision is most effective for texture-rich and semantics-heavy elements; LiDAR is critical for geometry- and boundary-oriented perception; radar can improve robustness under degraded visibility but is rarely sufficient for fine-grained infrastructure parsing; and GNSS/IMU, HD maps, and map priors mainly support ego-motion alignment, temporal consistency, map consistency, and structured reasoning. This evidence-level view explains why road infrastructure perception increasingly requires multi-modal, map-assisted, and representation-aware sensing pipelines~\cite{wang2023multimodal,cui2022deep}.

\subsection{Scene Representations for Road Infrastructure}

A defining characteristic of road infrastructure perception is that its outputs are more structured than conventional object-centric predictions. The goal is not only to localize individual elements, but also to recover road layout, functional driving space, regulatory meaning, and their spatial relations in forms usable for mapping and planning \cite{george2025infrastructure}. Fig.~\ref{fig:sensing_representation} illustrates this sensing-to-representation perspective across different infrastructure categories. Representation is therefore not a secondary implementation choice, but a central dimension of the field, because it determines how sensing evidence can be transformed into downstream-consumable road-scene structure~\cite{li2024fastbev}.

Earlier infrastructure perception methods commonly rely on dense raster representations, such as image-space masks, heatmaps, point-wise labels, or voxel-level predictions. These representations are easy to supervise and preserve local visual or geometric detail, making them suitable for lane segmentation, drivable-area estimation, curb segmentation, parking slot marking extraction, and road-marking parsing~\cite{muhammad2022vision}. However, they are often view-dependent, weak in topology preservation, and difficult to consume directly by map construction or motion planning without additional post-processing.

To improve spatial consistency, BEV representations have become increasingly central~\cite{li2024fastbev,li2024bevformer}. BEV provides a metric and planning-compatible canvas that reduces perspective distortion and organizes lanes, road boundaries, drivable areas, parking layouts, and map elements in a shared top-down frame. This makes BEV particularly effective for cross-modal fusion, multi-task infrastructure perception, and map-aligned spatial reasoning, although its quality depends on accurate projection, feature lifting, calibration, and resolution design.

Beyond rasterized BEV outputs, the field is moving toward structured vector and topology-aware representations. Vectorized lanes, road boundaries, parking slot primitives, and map elements provide compact and continuous geometric descriptions that are more compatible with HD maps, online map updating, and planning interfaces~\cite{liu2024monocular}. Graph-based representations further encode connectivity, adjacency, successor relations, merge/diverge structures, and lane-level topology, which are essential for intersection understanding and map-style reasoning. These representations offer stronger downstream utility, but also introduce more complex annotation, learning, and evaluation challenges.

For functional-spatial tasks, occupancy- and traversability-oriented representations provide a direct interface to motion planning by describing whether surrounding regions are accessible~\cite{xie2023circular}. For regulatory-semantic tasks, representation is expanding from bounding boxes and masks toward semantic-symbolic representations, where traffic signs, traffic lights, arrows, road text, stop lines, and crosswalks are linked to rule meaning, lane usage, and map-level semantics~\cite{yang2024traffic,guo2024signparser}. This trend is especially important for lane-level grounding, rule-aware reasoning, and explainable autonomous driving.

Overall, the evolution of road infrastructure perception is also an evolution of representation: from dense appearance-based prediction toward metric, structural, relational, and rule-aware scene modeling. Table~\ref{tab:infra_repr} summarizes the main representation families, their encoded structures, representative tasks, and downstream limitations.

\begin{table*}[t]
\renewcommand{\arraystretch}{1.12}
\caption{\textbf{Scene representation families for road infrastructure perception.}}
\centering
\setlength{\tabcolsep}{1.3mm}
\scriptsize
\resizebox{\textwidth}{!}{
\begin{tabular}{p{2.63cm} p{4.2cm} p{4.6cm} p{5.4cm}}
\toprule
\rowcolor{gray!10}
\textbf{Representation} 
& \textbf{Encoded Structure} 
& \textbf{Representative Tasks} 
& \textbf{Utility and Limitation} \\
\midrule

\multirow{2}{=}{Dense raster}
& Pixel-, point-, voxel-, or grid-level labels
& Lane detection; drivable-space; curb detection; road-marking recognition
& Fine-grained parsing, but view-dependent, topology-weak, and hard to use for mapping/planning \\

\rowcolor{gray!5}
\multirow{2}{=}{BEV grid}
& Metric top-down layout in ego/map-aligned coordinates
& Lane detection; drivable-space; parking slot detection; online map perception
& Metric reasoning and multi-modal fusion, but sensitive to projection, calibration, and grid resolution \\

\multirow{2}{=}{Structured vector}
& Polylines, splines, boundaries, polygons, and primitives
& Lane detection; curb detection; parking slot detection; map-element perception
& HD-map and planning compatibility, but sensitive to discontinuity, ordering, and matching errors \\

\rowcolor{gray!5}
\multirow{2}{=}{Topology-aware graph}
& Connectivity, adjacency, successor, merge/diverge, and lane-level relations
& Lane topology; intersection perception; lane-level map construction
& Routing and relation reasoning, but costly to annotate and difficult to evaluate \\

\multirow{2}{=}{Occupancy / traversability}
& Occupancy states, accessibility fields, traversability maps, and cost-map fields
& Drivable-space perception; free-space estimation; local planning
& Direct planning interface, but ambiguous under occlusion, temporary changes, and weak boundaries \\

\rowcolor{gray!5}
\multirow{2}{=}{Semantic-symbolic}
& Rule-bearing symbols, signal states, lane usage, attributes, and constraints
& Traffic sign recognition; traffic-light state; road-marking recognition
& Rule-compliant behavior, but requires grounding, lane association, and regional adaptation \\
\bottomrule
\end{tabular}}
\label{tab:infra_repr}
\vspace{-10pt}
\end{table*}

\section{Core Tasks in Road Infrastructure Perception} 
\label{sec:core_tasks}

\subsection{Overview of Task Organization}

Building upon the taxonomy and representation frameworks established in the previous sections, this section organizes core tasks in road infrastructure perception from a unified, problem-oriented perspective. Rather than treating each detection, segmentation, or recognition task in isolation, we view them as different instances of a shared objective: recovering structured, functionally meaningful, and regulation-aware representations of road environments for autonomous driving.

\begin{figure*}[t!]
    \centering
    \includegraphics[width=1.0\textwidth]{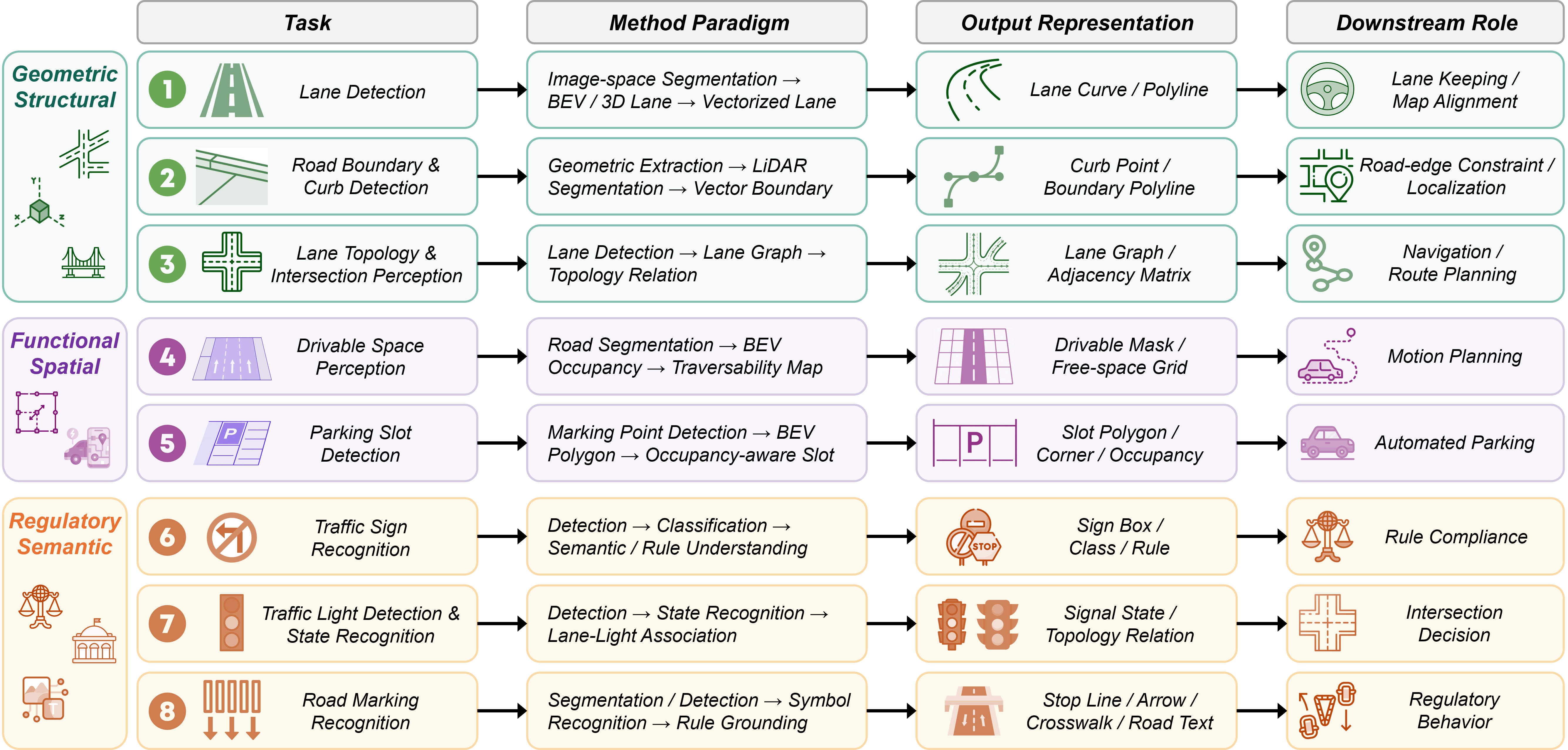}
    \vspace{-14pt}
    \caption{\textbf{Task-oriented method paradigms and output representations.}
Road infrastructure perception tasks differ in target structure and downstream role, while their methodological evolution shows a broad trend from image-space dense prediction toward BEV, vectorized, graph-based, occupancy-oriented, and semantic-symbolic representations.}
    \label{modeling-framework}
    \vspace{-15pt}
\end{figure*}

To enable consistent analysis across heterogeneous tasks, we characterize each task along four dimensions: (i) \emph{problem nature}, which reflects whether the task focuses on geometric reconstruction, spatial reasoning, or semantic interpretation; (ii) \emph{structural constraints}, which describe intrinsic properties of the target, such as continuity, boundary discontinuity, connectivity, region occupancy, or symbolic discreteness; (iii) \emph{representation tendency}, indicating preferred output forms including raster, BEV, vector, graph, occupancy, or semantic-symbolic representations; and (iv) \emph{functional role}, which captures how the task supports downstream autonomy modules such as localization, mapping, planning, parking, and rule-compliant decision-making. Fig.~\ref{modeling-framework} summarizes this view by linking each infrastructure task to its dominant modeling paradigm, output representation, and downstream role.

This unified view highlights both distinctions and dependencies among tasks. Lane and curb perception both address geometric structure, but differ in continuity modeling versus boundary inference. Drivable space and parking slot perception both concern functional accessibility, but operate at different spatial scales and maneuvering regimes. Traffic signs, traffic lights, and road markings all encode regulatory semantics, yet differ in whether the relevant information is static, dynamic, vertical, surface-embedded, or lane-associated. At the same time, these tasks are tightly coupled: lane structures support topology reasoning, road boundaries constrain drivable regions, parking slots define local maneuvering spaces, and regulatory elements such as signs, lights, and markings jointly determine rule-compliant behavior within lane-level contexts.

Based on this framework, we organize core tasks into geometric-structural, functional-spatial, and regulatory-semantic categories. Each task is discussed in terms of its problem formulation, structural properties, modeling paradigms, representation evolution, and key challenges, while the corresponding task-specific tables provide compact literature indexes for representative studies.

\subsection{Geometric-Structural Infrastructure Tasks}

\subsubsection{\textbf{Lane Detection}}
Lane detection aims to recover visible or inferable lane-line structures, including lane markings, lane dividers, lane curves, and lane instances, from onboard sensor observations. Within road infrastructure perception, it is a representative geometric-structural task because lane lines provide lane-level primitives that organize traffic flow, support lane keeping, constrain trajectory planning, and enable map alignment and HD map construction. Different from generic semantic segmentation, lane detection requires structured reasoning over thin, elongated, continuous, and often partially occluded line patterns whose appearance is strongly affected by perspective distortion, illumination changes, shadows, worn markings, and complex road curvature. Representative studies are summarized in Table~\ref{tab:lane_detection}, following the evolution from image-based 2D detection to BEV/3D estimation and vectorized map-compatible representation.

Early deep lane detection methods mainly formulate the task in the front-view image space. Segmentation-based and feature-propagation methods exploit spatial context to preserve the continuity of sparse lane markings, while row/column-wise classification, anchor-based detection, transformer attention, conditional instance decoding, and curve regression reduce dense pixel prediction into compact lane-instance or lane-shape estimation problems \cite{pan2018spatial,hou2019learning,qin2024ultrafast,zheng2021resa,li2024pganet,tabelini2021keep,zhao2024hwlane,liu2021condlanenet,zheng2025clrnetv2,feng2022rethinking,jin2022eigenlanes}. These image-based 2D methods are computationally efficient and effective for standard front-view benchmarks, but their outputs remain tied to perspective image coordinates. As a result, long-range lane geometry, lane-instance separation near vanishing regions, and occluded or weakly visible markings are difficult to interpret for planning or map-compatible reasoning.

To reduce perspective ambiguity, BEV and 3D lane detection methods explicitly recover lane geometry in BEV or 3D space. 3D-LaneNet \cite{garnett20193d} and Gen-LaneNet \cite{guo2020gen} introduce image-to-ground geometric transformations and 3D lane anchors, while later methods improve spatial feature lifting, BEV representation, and 3D query modeling through perspective transformers, virtual-camera normalization, 3D anchors, front-view 3D-aware attention, continuous spline modeling, or frequency-aware feature fusion \cite{garnett20193d,guo2020gen,chen2022persformer,yan2022once,wang2023bev,huang2023anchor3dlane,luo2023latr,pittner2024lanecpp,song2025freq3dlane}. This evolution reflects a fundamental requirement of autonomous driving: lane detections should not only be visually accurate in the image plane, but also geometrically meaningful in the vehicle coordinate system, especially for slopes, bumps, long-range planning, and lane-level localization.

More recently, vectorized and map-compatible formulations represent lanes as polylines, splines, structured curves, point sets, or lane graphs. Such representations bridge lane detection and online map construction by producing compact geometric elements that can be integrated into local maps, topology reasoning modules, and planning interfaces \cite{can2021structured,li2022hdmapnet,liu2023vectormapnet,liao2023maptr,liao2024lane,li2026unified}. Compared with raster masks, vector lanes better preserve continuity, directionality, and connectivity, which are essential for downstream lane topology reasoning and map updates. However, they also require robust instance ordering, topology recovery, and temporal consistency under occlusion and sparse supervision.

Overall, lane detection has shifted from image-space visual localization toward metrically grounded and map-compatible lane representation. Despite substantial progress, open challenges remain in detecting faded or occluded markings, handling adverse weather and night scenes, generalizing across regions with different lane-painting conventions, maintaining temporal stability, and producing lane representations that are simultaneously accurate, continuous, topologically valid, and directly usable by planning and mapping systems.

\begin{table*}[t]
\renewcommand\arraystretch{1.25}
\caption{\textbf{Representative studies on lane detection.}
The input icons indicate the input modalities used by each method:
\includegraphics[width=0.018\textwidth]{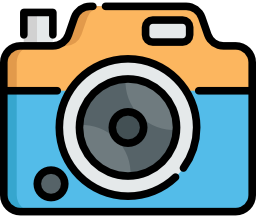} denotes camera input, and
\includegraphics[width=0.018\textwidth]{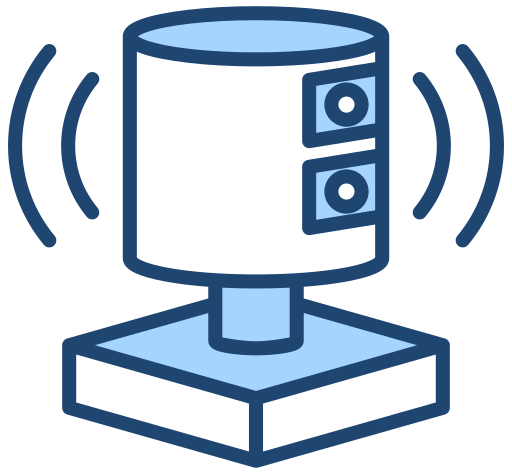} denotes LiDAR input.
}
\centering
\scriptsize
\setlength{\tabcolsep}{1.2mm}
\resizebox{\textwidth}{!}{
\begin{tabular}{p{1.9cm} p{1.cm} C{1.4cm} p{1.6cm} p{2.3cm} p{3.8cm} p{7.6cm}}
\hline
\rowcolor{gray!10}
\textbf{Method} & \textbf{Venue} & \textbf{Input} & \textbf{Input Type} & \textbf{Output} & \textbf{Dataset} & \textbf{Key idea} \\
\hline

\rowcolor{tpami_yellow!10}
\multicolumn{7}{l}{\textit{\textcolor{black}{Image-based 2D Lane Detection}}} \\
Pan et al. \cite{pan2018spatial} & AAAI'18 & \includegraphics[width=0.18\linewidth]{fig/icon/rgb.png} & Monocular & 2D Lane Segmentation & CULane, TuSimple & Spatial message passing for long and thin lane structures. \\
Hou et al. \cite{hou2019learning} & ICCV'19 & \includegraphics[width=0.18\linewidth]{fig/icon/rgb.png} & Monocular & 2D Lane Segmentation & CULane, TuSimple & Self-attention distillation for lightweight lane segmentation. \\
Qin et al. \cite{qin2024ultrafast} & TPAMI'24 & \includegraphics[width=0.18\linewidth]{fig/icon/rgb.png} & Monocular & 2D Lane Points & CULane, TuSimple, CurveLanes, LLAMAS & Hybrid row-column anchors with ordinal classification for fast lane detection. \\
Zheng et al. \cite{zheng2021resa} & AAAI'21 & \includegraphics[width=0.18\linewidth]{fig/icon/rgb.png} & Monocular & 2D Lane Segmentation & CULane, TuSimple & Recurrent feature-shift aggregation for lane continuity modeling. \\
Li et al. \cite{li2024pganet} & T-ITS'24 & \includegraphics[width=0.18\linewidth]{fig/icon/rgb.png} & Monocular & 2D Lane Curves & TuSimple, CULane, LLAMAS & Polynomial global attention and curvature loss for lane modeling. \\
Tabelini et al. \cite{tabelini2021keep} & CVPR'21 & \includegraphics[width=0.18\linewidth]{fig/icon/rgb.png} & Monocular & 2D Lane Anchors & CULane, TuSimple, LLAMAS & Anchor attention pooling for real-time lane detection. \\
Zhao et al. \cite{zhao2024hwlane} & T-ITS'24 & \includegraphics[width=0.18\linewidth]{fig/icon/rgb.png} & Monocular & 2D Lane Points & CULane, TuSimple, LLAMAS & Row- and column-wise transformer attention for lane structure modeling. \\
Liu et al. \cite{liu2021condlanenet} & ICCV'21 & \includegraphics[width=0.18\linewidth]{fig/icon/rgb.png} & Monocular & 2D Lane Points & CULane, TuSimple, CurveLanes & Conditional convolution for instance-specific lane decoding. \\
Zheng et al. \cite{zheng2025clrnetv2} & TPAMI'25 & \includegraphics[width=0.18\linewidth]{fig/icon/rgb.png} & Monocular & 2D Lane Curves & CULane, TuSimple, CurveLanes & Cross-layer refinement with dense-lane discrimination and Line regression. \\
Feng et al. \cite{feng2022rethinking} & CVPR'22 & \includegraphics[width=0.18\linewidth]{fig/icon/rgb.png} & Monocular & 2D Lane Curves & CULane, TuSimple, LLAMAS & End-to-end parametric B\'ezier curve modeling for efficient lane detection. \\
Jin et al. \cite{jin2022eigenlanes} & CVPR'22 & \includegraphics[width=0.18\linewidth]{fig/icon/rgb.png} & Monocular & 2D Lane Curves & TuSimple, CULane, SDLane & Low-rank data-driven lane shape descriptors for diverse lane structures. \\
Feng et al. \cite{feng2025elasticlanenet} & WACV'25 & \includegraphics[width=0.18\linewidth]{fig/icon/rgb.png} & Monocular & 2D Lane Curves & SDLane, TuSimple, CULane & Elastic lane map for geometry-flexible lane modeling. \\
\hline

\rowcolor{tpami_yellow!10}
\multicolumn{7}{l}{\textit{\textcolor{black}{BEV and 3D Lane Detection}}} \\
Garnett et al. \cite{garnett20193d} & ICCV'19 & \includegraphics[width=0.18\linewidth]{fig/icon/rgb.png} & Monocular & 3D Lane Anchors & 3D-Lanes & Intra-network inverse perspective mapping for monocular 3D lane detection. \\
Guo et al. \cite{guo2020gen} & ECCV'20 & \includegraphics[width=0.18\linewidth]{fig/icon/rgb.png} & Monocular & 3D Lane Points & Apollo & Geometry-guided anchor representation for scalable 3D lane detection. \\
Chen et al. \cite{chen2022persformer} & ECCV'22 & \includegraphics[width=0.18\linewidth]{fig/icon/rgb.png} & Monocular & 3D Lane Anchors & OpenLane, Apollo & Perspective transformer for lifting image features into 3D lane space. \\
Yan et al. \cite{yan2022once} & CVPR'22 & \includegraphics[width=0.18\linewidth]{fig/icon/rgb.png} & Monocular & 3D Lane Points & ONCE-3DLanes & Extrinsic-free spatial regression benchmark and baseline for 3D lanes. \\
Wang et al. \cite{wang2023bev} & CVPR'23 & \includegraphics[width=0.18\linewidth]{fig/icon/rgb.png} & Monocular & BEV Lane Points & OpenLane, Apollo & Virtual camera transformation with BEV keypoint representation. \\
Huang et al. \cite{huang2023anchor3dlane} & CVPR'23 & \includegraphics[width=0.18\linewidth]{fig/icon/rgb.png} & Monocular & 3D Lane Anchors & OpenLane, Apollo, ONCE & BEV-free 3D anchor regression from front-view image features. \\
Luo et al. \cite{luo2023latr} & ICCV'23 & \includegraphics[width=0.18\linewidth]{fig/icon/rgb.png} & Monocular & 3D Lane Curves & OpenLane, ONCE, Apollo & Lane-aware transformer queries with dynamic 3D positional embedding. \\
Pittner et al. \cite{pittner2024lanecpp} & CVPR'24 & \includegraphics[width=0.18\linewidth]{fig/icon/rgb.png} & Monocular & 3D Lane Curves & OpenLane, Apollo, ONCE & Continuous spline-based 3D lane representation with physical priors. \\
Song et al. \cite{song2025freq3dlane} & T-ITS'25 & \includegraphics[width=0.18\linewidth]{fig/icon/rgb.png} & Monocular & 3D Lane Points & OpenLane, Apollo & Frequency-aware feature fusion for monocular 3D lane detection. \\
\hline

\rowcolor{tpami_yellow!10}
\multicolumn{7}{l}{\textit{\textcolor{black}{Vectorized and Map-compatible Lane Detection}}} \\
Can et al. \cite{can2021structured} & ICCV'21 & \includegraphics[width=0.18\linewidth]{fig/icon/rgb.png} & Monocular & BEV Lane Graph & nuScenes, Argoverse & Structured BEV lane graph extraction from onboard images. \\
Li et al. \cite{li2022hdmapnet} & ICRA'22 & \includegraphics[width=0.18\linewidth]{fig/icon/rgb.png}~\includegraphics[width=0.18\linewidth]{fig/icon/lidar.png} & Surround-view/LiDAR & Vectorized Map & nuScenes & Online BEV semantic map construction with lane-level elements. \\
Liu et al. \cite{liu2023vectormapnet} & ICML'23 & \includegraphics[width=0.18\linewidth]{fig/icon/rgb.png}~\includegraphics[width=0.18\linewidth]{fig/icon/lidar.png} & Surround-view/LiDAR & Map Polylines & nuScenes, Argoverse 2 & End-to-end sparse vectorized HD map learning. \\
Liao et al. \cite{liao2023maptr} & ICLR'23 & \includegraphics[width=0.18\linewidth]{fig/icon/rgb.png} & Surround-view & Vectorized Map & nuScenes & Permutation-equivalent structured modeling for vectorized map construction. \\
Liao et al. \cite{liao2024lane} & ECCV'24 & \includegraphics[width=0.18\linewidth]{fig/icon/rgb.png} & Surround-view & Lane Graph & nuScenes, Argoverse 2, OpenLane-V2 & Path-wise continuity-preserving lane graph construction. \\
Li et al. \cite{li2026unified} & TCSVT'26 & \includegraphics[width=0.18\linewidth]{fig/icon/rgb.png} & Surround-view & Lane Graph & OpenLane-V2 & Connected-lane queries for joint topology reasoning. \\
\hline
\end{tabular}}
\vspace{-12pt}
\label{tab:lane_detection}
\end{table*}

\subsubsection{\textbf{Road Boundary and Curb Detection}}

Road boundary and curb detection aims to recover the physical limits of the roadway, including curbs, road edges, shoulders, elevation discontinuities, medians, and unmarked structural boundaries. Unlike lane detection, which mainly models semantic centerlines or painted lane markings, this task is fundamentally concerned with geometric-structural boundary inference. Curbs and road boundaries delimit the transition between drivable and non-drivable surfaces, and provide critical constraints for drivable-space estimation, free-space reasoning, lateral localization, HD map construction, and planning in weakly marked or unmarked environments \cite{hata2016feature,zhang2018road,xu2021topo,zhao2025curbnet}. In this sense, road boundaries connect low-level geometric perception with infrastructure-level scene organization. Table~\ref{tab:road_boundary_curb} summarizes representative studies from geometric extraction to learned, LiDAR-based, and map-compatible boundary representations.

The structural nature of this task makes it more geometry-dependent than conventional appearance-based road parsing. Lane markings are often visually salient but may disappear due to occlusion, wear, snow, or regional variation; by contrast, curbs and road edges are often defined by height jumps, slope changes, surface discontinuities, and long-range spatial continuity. Road boundaries are also closely related to, but not identical with, lane boundaries and drivable-space limits. Lane boundaries describe lane-level organization inside the road surface, whereas road boundaries define the outer physical envelope of the road. Drivable-space segmentation provides an area-level mask, while curb and boundary detection seeks a compact structural representation, such as boundary points, polylines, BEV contours, or map elements.

Early studies mainly followed classical geometric extraction pipelines, where candidate curb or road-edge points were identified from LiDAR scan lines, elevation gradients, local saliency, supervoxels, graph cuts, or curve fitting \cite{hata2016feature,xu2017road,zai2018road,zhang2018road,mi2022automated}. These methods are interpretable and metric-accurate, but they often rely on hand-crafted thresholds and degrade under occlusion, sparse sensing, irregular curb shapes, or complex intersections. Image-based 2D approaches instead formulate road boundary perception as visible or occluded boundary segmentation, monocular curb localization, or aerial-image graph extraction \cite{suleymanov2018inferring,panev2019road,xu2021icurb,xu2021cp,xu2021topo,xu2022csboundary}. They improve semantic context modeling and can exploit large-scale aerial imagery for offline HD map annotation, but perspective ambiguity and weak geometric observability remain limiting factors.

LiDAR-based 3D detection has become important because point clouds directly encode the metric height and spatial discontinuities that define curbs. Recent methods project sparse LiDAR into BEV representations, learn uncertainty-aware curb segmentation, construct curb-specific datasets, or design point-cloud segmentation frameworks that emphasize height-sensitive features \cite{suleymanov2019online,jung2021uncertainty,bai2022build,zhao2025curbnet}. Meanwhile, map-based and multi-modal methods push the task beyond local edge extraction toward structured, map-compatible representation. Road boundaries are increasingly modeled as polylines or vectorized map elements using camera-LiDAR fusion, satellite imagery, BEV maps, and online HD map construction frameworks \cite{liang2019convolutional,ma2022boundarynet,xiong2022road,li2022hdmapnet,liu2023vectormapnet,liao2023maptr,ding2023pivotnet,yuan2024streammapnet}.

Overall, road boundary and curb detection is moving from handcrafted geometric extraction toward learned, vectorized, and map-compatible boundary representation. Remaining challenges include robust perception under occlusion and adverse weather, continuity preservation at intersections, cross-region generalization, sparse long-range boundary recovery, and unified evaluation across point, mask, polyline, BEV-contour, and graph representations.

\begin{table*}[t]
\renewcommand\arraystretch{1.12}
\caption{\textbf{Representative studies on road boundary and curb detection.}
The input icons indicate the input modalities used by each method:
\includegraphics[width=0.018\textwidth]{fig/icon/rgb.png} denotes camera input,
\includegraphics[width=0.018\textwidth]{fig/icon/lidar.png} denotes LiDAR input, and
\includegraphics[width=0.018\textwidth]{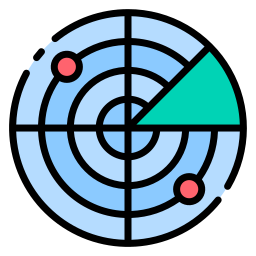} denotes radar input.
}
\centering
\scriptsize
\setlength{\tabcolsep}{1.1mm}
\resizebox{\textwidth}{!}{
\begin{tabular}{p{2.3cm} p{1.2cm} C{1.4cm} p{2.4cm} p{2.5cm} p{2.5cm} p{6.cm}}
\hline
\rowcolor{gray!10}
\textbf{Method} & \textbf{Venue} & \textbf{Input} & \textbf{Input Type} & \textbf{Output} & \textbf{Dataset} & \textbf{Key Idea} \\
\hline

% ==================== Classical Geometric Extraction ====================
\rowcolor{tpami_yellow!10}
\multicolumn{7}{l}{\textit{Classical Geometric Extraction}} \\
Hata et al. \cite{hata2016feature} & T-ITS'16 & \includegraphics[width=0.18\linewidth]{fig/icon/lidar.png} & LiDAR & Curb Points & Private & Extracts curb and marking features for LiDAR localization. \\
Xu et al. \cite{xu2017road} & T-GRS'17 & \includegraphics[width=0.18\linewidth]{fig/icon/lidar.png} & LiDAR & Curb Points & Private & Extracts curb candidates with energy-based refinement. \\
Zai et al. \cite{zai2018road} & T-ITS'18 & \includegraphics[width=0.18\linewidth]{fig/icon/lidar.png} & LiDAR & Boundary Points & Private & Extracts boundaries using supervoxels and graph cuts. \\
Zhang et al. \cite{zhang2018road} & T-ITS'18 & \includegraphics[width=0.18\linewidth]{fig/icon/lidar.png} & LiDAR & Curb Points & Private & Extracts curbs after road-surface segmentation. \\
Mi et al. \cite{mi2022automated} & T-ITS'22 & \includegraphics[width=0.18\linewidth]{fig/icon/lidar.png} & LiDAR & Boundary Polylines & Private & Vectorizes road boundaries from LiDAR scans. \\
Xu et al. \cite{xu2025urban} & Measurement'25 & \includegraphics[width=0.18\linewidth]{fig/icon/lidar.png} & LiDAR & Boundary Points & SemanticKITTI & Classifies road shapes for boundary extraction. \\
\hline

% ==================== Image-based 2D Detection ====================
\rowcolor{tpami_yellow!10}
\multicolumn{7}{l}{\textit{Image-based 2D Detection}} \\
Suleymanov et al. \cite{suleymanov2018inferring} & ITSC'18 & \includegraphics[width=0.18\linewidth]{fig/icon/rgb.png} & Monocular & Boundary Segmentation & Oxford RobotCar & Infers visible and occluded road boundaries. \\
Panev et al. \cite{panev2019road} & T-ITS'19 & \includegraphics[width=0.18\linewidth]{fig/icon/rgb.png} & Fisheye & Boundary Polylines & Private & Estimates curb geometry from fisheye imagery. \\
Xu et al. \cite{xu2021icurb} & RA-L'21 & \includegraphics[width=0.18\linewidth]{fig/icon/rgb.png} & Aerial-view & Boundary Graph & Topo-Boundary & Traces boundary graphs with imitation learning. \\
Xu et al. \cite{xu2021cp} & IROS'21 & \includegraphics[width=0.18\linewidth]{fig/icon/rgb.png} & Aerial-view & Boundary Segmentation & Topo-Boundary & Preserves connectivity in boundary segmentation. \\
Xu et al. \cite{xu2021topo} & RA-L'21 & \includegraphics[width=0.18\linewidth]{fig/icon/rgb.png} & Aerial-view & Boundary Graph & Topo-Boundary & Introduces topological boundary benchmark and metrics. \\
Xu et al. \cite{xu2022csboundary} & RA-L'22 & \includegraphics[width=0.18\linewidth]{fig/icon/rgb.png} & Aerial-view & Boundary Graph & Topo-Boundary & Stitches city-scale boundary graphs. \\
\hline

% ==================== LiDAR-based 3D Detection ====================
\rowcolor{tpami_yellow!10}
\multicolumn{7}{l}{\textit{LiDAR-based 3D Detection}} \\
Suleymanov et al. \cite{suleymanov2019online} & ITSC'19 & \includegraphics[width=0.18\linewidth]{fig/icon/lidar.png} & LiDAR & Curb Segmentation & Oxford RobotCar & Infers curbs from sparse LiDAR. \\
Jung et al. \cite{jung2021uncertainty} & ICRA'21 & \includegraphics[width=0.18\linewidth]{fig/icon/lidar.png} & LiDAR & Curb Segmentation & Private & Adds uncertainty to LiDAR curb segmentation. \\
Bai et al. \cite{bai2022build} & ICRA'22 & \includegraphics[width=0.18\linewidth]{fig/icon/lidar.png} & LiDAR & Curb Segmentation & SemanticKITTI & Generates curb labels from LiDAR sequences. \\
Sha et al. \cite{sha2020supervoxel} & GRSL'22 & \includegraphics[width=0.18\linewidth]{fig/icon/lidar.png} & LiDAR & Boundary Points & IQmulus/TerraMobilita, Semantic3D & Enhances boundaries with supervoxel structures. \\
Zhao et al. \cite{zhao2025curbnet} & T-ITS'25 & \includegraphics[width=0.18\linewidth]{fig/icon/lidar.png} & LiDAR & Curb Segmentation & NRS, 3D-Curb & Segments curbs using height-aware features. \\
\hline

% ==================== Map-based and Multi-modal Detection ====================
\rowcolor{tpami_yellow!10}
\multicolumn{7}{l}{\textit{Map-based and Multi-modal Detection}} \\
Liang et al. \cite{liang2019convolutional} & CVPR'19 & \includegraphics[width=0.18\linewidth]{fig/icon/rgb.png}~\includegraphics[width=0.18\linewidth]{fig/icon/lidar.png} & Monocular/LiDAR & Boundary Curves & Private & Extracts polylines with convolutional recurrence. \\
Ma et al. \cite{ma2022boundarynet} & T-ITS'22 & \includegraphics[width=0.18\linewidth]{fig/icon/rgb.png}~\includegraphics[width=0.18\linewidth]{fig/icon/lidar.png} & Satellite/LiDAR & Boundary Curves & Paris-Lille-3D & Completes boundaries with LiDAR and satellite imagery. \\
Xiong et al. \cite{xiong2022road} & T-ITS'22 & \includegraphics[width=0.18\linewidth]{fig/icon/lidar.png} & LiDAR & HD Map Boundary & Private & Extracts map boundaries using road models. \\
Li et al. \cite{li2022hdmapnet} & ICRA'22 & \includegraphics[width=0.18\linewidth]{fig/icon/rgb.png}~\includegraphics[width=0.18\linewidth]{fig/icon/lidar.png} & Surround-view/LiDAR & Vectorized Map & nuScenes & Constructs local HD maps from sensors. \\
Liu et al. \cite{liu2023vectormapnet} & ICML'23 & \includegraphics[width=0.18\linewidth]{fig/icon/rgb.png}~\includegraphics[width=0.18\linewidth]{fig/icon/lidar.png} & Surround-view/LiDAR & Vectorized Map & nuScenes, Argoverse 2 & Predicts BEV map polylines end-to-end. \\
Liao et al. \cite{liao2023maptr} & ICLR'23 & \includegraphics[width=0.18\linewidth]{fig/icon/rgb.png} & Surround-view & Vectorized Map & nuScenes & Represents maps as structured point sets. \\
Ding et al. \cite{ding2023pivotnet} & ICCV'23 & \includegraphics[width=0.18\linewidth]{fig/icon/rgb.png} & Surround-view & Vectorized Map & nuScenes, Argoverse 2 & Learns compact pivots for map elements. \\
Yuan et al. \cite{yuan2024streammapnet} & WACV'24 & \includegraphics[width=0.18\linewidth]{fig/icon/rgb.png} & Surround-view & Vectorized Map & nuScenes, Argoverse 2 & Uses temporal streaming for stable mapping. \\
Kingery et al. \cite{kingery2024road} & IROS'24 & \includegraphics[width=0.18\linewidth]{fig/icon/radar.png} & Radar & Boundary Curves & private & Estimates boundaries from sparse radar. \\
\hline
\end{tabular}
}
\vspace{-12pt}
\label{tab:road_boundary_curb}
\end{table*}

\subsubsection{\textbf{Lane Topology and Intersection Perception}}

Lane topology and intersection perception aims to recover the relational organization of road lanes rather than only estimating lane markings or lane geometries. In contrast to lane detection, which primarily outputs lane-line curves, lane instances, or centerline primitives, this task focuses on how lanes are connected, ordered, merged, split, and constrained by traffic flow. Its output is therefore a structured infrastructure representation, such as a directed lane graph, topology matrix, adjacency matrix, lane connector set, or intersection-level layout. Such representations are essential for lane-level navigation, route planning, behavior prediction, online map updating, and decision-making in complex urban scenarios, where the key issue is not only where lanes are located but also how vehicles can move among them \cite{homayounfar2019dagmapper,zurn2021lane,can2021structured,can2022topology}. Table~\ref{tab:lane_topology_intersection} summarizes representative studies on topology-aware lane graph perception and intersection layout perception.

Existing works can be summarized into two closely related paradigms. The first is topology-aware lane graph perception, where lane elements are represented as graph nodes, polylines, lane segments, paths, or topology-aware BEV structures. Early methods formulate lane topology as directed graphical inference, lane-anchor graph estimation, or structured BEV road-graph extraction, enabling explicit modeling of forks, merges, lane directions, and successor relations \cite{homayounfar2019dagmapper,zurn2021lane,can2021structured}. Later works further improve topological consistency by introducing minimal-cycle supervision, global lane graph aggregation, routable aerial lane maps, and path-wise graph modeling \cite{can2022topology,he2022lane,buchner2023learning,liao2024lane}. With the release of topology reasoning benchmarks, recent methods increasingly model lane-lane and lane-traffic relationships using scene graphs, relation heads, topology matrices, sequence encoding, and endpoint-aware reasoning \cite{wang2023openlanev2,li2023toponet,wu2024topomlp,peng2024lanegraph2seq,fu2026topopoint}. These methods indicate that reliable lane topology perception requires both accurate geometry and structurally consistent connectivity.

The second paradigm is intersection layout perception, which focuses on lane-level organization inside junctions. Intersections are more difficult than regular road segments because lane markings may be incomplete, connectors can be virtual, and multiple inbound lanes may correspond to different outbound lanes through left-turn, right-turn, straight-through, merge, and diverge relations. Representative studies infer lane-based intersection maps from crowdsourced traces, estimate lane-level intersection topology from surrounding-agent trajectories, calibrate missing or incorrect turning paths from large-scale trajectory data, and predict lane-to-lane adjacency matrices from aerial imagery and OpenStreetMap priors \cite{yang2018generating,meyer2019anytime,meyer2020fast,zhao2020automatic,nilsson2026predicting}. Other works exploit roadside LiDAR, traffic flow observations, or semantic-particle filtering to reconstruct intersection geometry, lane directions, signal phases, virtual connectors, and intersection-level lane topology \cite{zhou2021automatic,qin2023traffic,liu2023automatic}. These methods show that intersection perception is not a simple extension of lane detection, but a structural reasoning task recovering hidden connectors and maneuver-level constraints.

Compared with lane detection, lane topology and intersection perception is more sensitive to structural errors: a geometrically plausible lane may still be invalid if connectivity, directionality, or maneuver constraints are wrong. Open challenges include robust endpoint association, topology recovery under occlusion or missing markings, generalization across cities and traffic rules, long-tail junction layouts, and unified evaluation that jointly considers geometry, connectivity, legality, and planning usability.

\begin{table*}[t]
\renewcommand\arraystretch{1.12}
\caption{\textbf{Representative studies on lane topology and intersection perception.}
The input icons indicate the input modalities or priors used by each method:
\includegraphics[width=0.018\textwidth]{fig/icon/rgb.png} denotes camera input,
\includegraphics[width=0.018\textwidth]{fig/icon/lidar.png} denotes LiDAR input,
\includegraphics[width=0.018\textwidth]{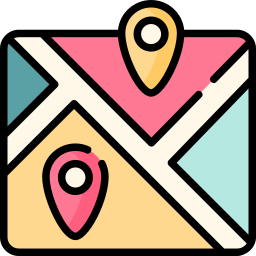} denotes HD map input,
\includegraphics[width=0.018\textwidth]{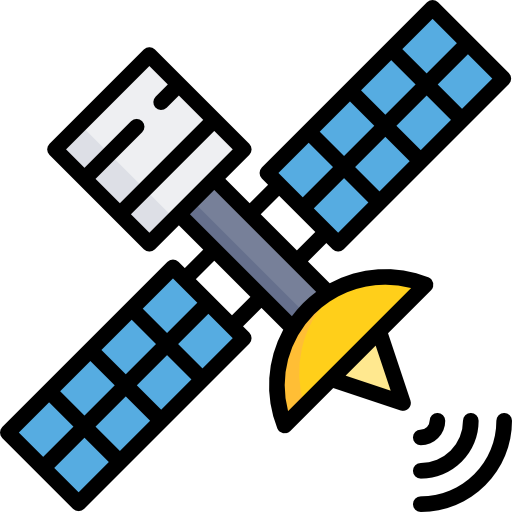} denotes GNSS input, and
\includegraphics[width=0.018\textwidth]{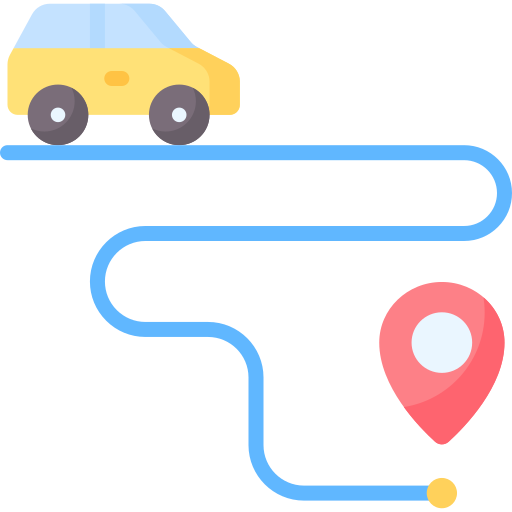} denotes trajectory input.
}
\centering
\scriptsize
\setlength{\tabcolsep}{1.2mm}
\resizebox{\textwidth}{!}{
\begin{tabular}{p{2.6cm} p{1.cm} C{1.4cm} p{2.cm} p{2.4cm} p{2.4cm} p{6.1cm}}
\hline
\rowcolor{gray!10}
\textbf{Method} & \textbf{Venue} & \textbf{Input} & \textbf{Input Type} & \textbf{Output} & \textbf{Dataset} & \textbf{Key Idea} \\
\hline

\rowcolor{tpami_yellow!10}
\multicolumn{7}{l}{\textit{Topology-aware Lane Graph Perception}} \\
Homayounfar et al. \cite{homayounfar2019dagmapper} & ICCV'19 & \includegraphics[width=0.18\linewidth]{fig/icon/lidar.png} & LiDAR & Lane Graph & Private & Traces lane-boundary topology from BEV LiDAR. \\
Z{\"u}rn et al. \cite{zurn2021lane} & RA-L'21 &
\includegraphics[width=0.18\linewidth]{fig/icon/rgb.png}
\includegraphics[width=0.18\linewidth]{fig/icon/lidar.png}
\includegraphics[width=0.18\linewidth]{fig/icon/map.png}
& Multi-modal BEV & Lane Graph & nuScenes &
Infers directed lane graphs from multi-modal BEV rasters. \\
Can et al. \cite{can2021structured} & ICCV'21 & \includegraphics[width=0.18\linewidth]{fig/icon/rgb.png} & Monocular & Lane Graph & nuScenes, Argoverse & Extracts BEV road graphs from onboard images. \\
Can et al. \cite{can2022topology} & CVPR'22 & \includegraphics[width=0.18\linewidth]{fig/icon/rgb.png} & Monocular & Lane Graph & nuScenes, Argoverse & Preserves topology with directed curves and minimal cycles. \\
He et al. \cite{he2022lane} & WACV'22 & \includegraphics[width=0.18\linewidth]{fig/icon/rgb.png} & Aerial-view & Lane Graph & Private & Extracts routable lane-level maps from aerial imagery. \\
B{\"u}chner et al. \cite{buchner2023learning} & CVPR'23 & \includegraphics[width=0.18\linewidth]{fig/icon/rgb.png} & Aerial-view & Lane Graph & UrbanLaneGraph & Aggregates local aerial lane graphs into global topology. \\
Wang et al. \cite{wang2023openlanev2} & NeurIPS'23 & \includegraphics[width=0.18\linewidth]{fig/icon/rgb.png} & Surround-view & Topology Graph & OpenLane-V2 & Benchmarks lane and traffic-element topology reasoning. \\
Wu et al. \cite{wu2024topomlp} & ICLR'24 & \includegraphics[width=0.18\linewidth]{fig/icon/rgb.png} & Surround-view & Topology Graph & OpenLane-V2 & Uses detector-first MLP heads for topology reasoning. \\
Liao et al. \cite{liao2024lane} & ECCV'24 & \includegraphics[width=0.18\linewidth]{fig/icon/rgb.png} & Surround-view & Lane Graph & nuScenes, Argoverse 2 & Models lane graphs as continuous traffic-flow paths. \\
Peng et al. \cite{peng2024lanegraph2seq} & AAAI'24 & \includegraphics[width=0.18\linewidth]{fig/icon/rgb.png} & Surround-view & Lane Graph & nuScenes, Argoverse 2 & Serializes vertices and edges for lane graph extraction. \\
Fu et al. \cite{fu2026topopoint} & NeurIPS'25 & \includegraphics[width=0.18\linewidth]{fig/icon/rgb.png} & Surround-view & Topology Graph & OpenLane-V2 & Refines topology with endpoint and point-lane reasoning. \\
Li et al. \cite{li2023toponet} & SCIS'26 & \includegraphics[width=0.18\linewidth]{fig/icon/rgb.png} & Surround-view & Topology Graph & OpenLane-V2 & Reasons lane-lane and lane-traffic topology with graphs. \\
\hline

\rowcolor{tpami_yellow!10}
\multicolumn{7}{l}{\textit{Intersection Layout Perception}} \\
Yang et al. \cite{yang2018generating} & TR-C'18 & \includegraphics[width=0.18\linewidth]{fig/icon/gnss.png} & GNSS Trajectory & Intersection Graph & Private & Generates lane-based intersection maps from traces. \\
Meyer et al. \cite{meyer2019anytime} & ITSC'19 & \includegraphics[width=0.18\linewidth]{fig/icon/traj.png} & Vehicle Trajectory & Intersection Graph & Private & Estimates intersection topology from participant trajectories. \\
Zhao et al. \cite{zhao2020automatic} & ICDE'20 & \includegraphics[width=0.18\linewidth]{fig/icon/traj.png} & Vehicle Trajectory & Intersection Graph & Private & Calibrates intersection topology from vehicle trajectories. \\
Meyer et al. \cite{meyer2020fast} & IV'20 & \includegraphics[width=0.18\linewidth]{fig/icon/traj.png} & Vehicle Trajectory & Intersection Graph & Private & Refines intersection lanes with MCMC and B-splines. \\
Zhou et al. \cite{zhou2021automatic} & IROS'21 & \includegraphics[width=0.18\linewidth]{fig/icon/rgb.png}~\includegraphics[width=0.18\linewidth]{fig/icon/map.png} & Monocular/Map & Vectorized Map & Private & Combines image semantics with OpenStreetMap priors. \\
Qin et al. \cite{qin2023traffic} & RA-L'23 & \includegraphics[width=0.18\linewidth]{fig/icon/traj.png} & Traffic Flow & Intersection Graph & Private & Infers complex topology from crowdsourced traffic flow. \\
Liu et al. \cite{liu2023automatic} & JAS'23 & \includegraphics[width=0.18\linewidth]{fig/icon/lidar.png} & LiDAR & Intersection Graph & Private & Generates intersection maps from roadside LiDAR tracks. \\
Nilsson et al. \cite{nilsson2026predicting} & TRIP'26 & \includegraphics[width=0.18\linewidth]{fig/icon/rgb.png}~\includegraphics[width=0.18\linewidth]{fig/icon/map.png} & Aerial-view/Map & Adjacency Matrix & Private & Predicts lane connectivity from aerial imagery and OSM. \\
\hline
\end{tabular}}
\vspace{-12pt}
\label{tab:lane_topology_intersection}
\end{table*}

\subsection{Functional-Spatial Infrastructure Tasks}

\subsubsection{\textbf{Drivable Space Perception}}
Drivable space perception estimates the spatial regions in which an autonomous vehicle can physically and operationally move. Unlike lane detection, which focuses on lane-line geometry, or road-boundary and curb detection, which localizes physical road limits, drivable space perception concerns functional accessibility. A region is drivable only when it is sufficiently traversable, obstacle-free, geometrically feasible, and compatible with near-term vehicle motion. Therefore, this task should not be reduced to ordinary road-surface segmentation: a road-like region may be non-drivable due to obstacles, traffic islands, temporary construction, occlusion, disconnected topology, or motion constraints. From the perspective of road infrastructure perception, drivable space forms a functional-spatial layer that links scene understanding with collision avoidance, local planning, emergency maneuvering, and map-free navigation. Table~\ref{tab:drivable_space} summarizes representative studies from image-space segmentation to BEV/occupancy and multi-modal free-space perception.

Early studies commonly formulate drivable space as image-space road or free-space segmentation. These approaches predict pixel-level road regions, drivable masks, or collision-free areas from monocular, stereo, or RGB-D images. Representative works exploit road appearance, traffic-object context, depth cues, edge constraints, or task-oriented pre-training to improve drivable-region estimation in urban driving scenes \cite{teichmann2018multinet,qian2020dlt,wang2021dynamic,han2024epurate,sheng2024edgeunet,ma2025task}. However, image-plane masks are perspective-dependent and do not directly encode metric accessibility, which limits their use in trajectory planning and cost-map construction.

To obtain planning-compatible representations, BEV and occupancy-based methods map drivable space into ego-centric top-down grids, semantic occupancy maps, or road-layout representations. In these representations, each cell corresponds to a physical region around the vehicle, facilitating reasoning about accessible road space, obstacle occupancy, and non-drivable regions. Monocular semantic occupancy mapping and BEV layout estimation predict road or free-space occupancy beyond visible image evidence \cite{lu2019monocular,mani2020monolayout}. Cross-view road-layout estimation methods further transform perspective images into BEV maps that jointly encode road layout and vehicle occupancy, providing a more spatially grounded representation for functional driving-space inference \cite{yang2021projecting,liu2024monocular}. LiDAR-based BEV methods directly estimate free space or drivable regions from metric sensing, improving robustness to illumination and appearance ambiguity \cite{popov2023nvradarnet,wen2024upbev}.

A further line of work emphasizes 3D and multi-modal free-space perception. RGB-D, stereo, LiDAR, radar, and camera--LiDAR fusion provide depth, surface-normal, ground-obstacle separation, and geometric consistency cues, which are crucial under weak texture, illumination changes, and partial occlusion. Surface-normal-enhanced freespace segmentation, depth-normal translation, cascaded LiDAR-camera fusion, LiDAR-only drivable-region detection, uncertainty-aware RGB-D fusion, stereo collision-free-space estimation, LiDAR-depth self-supervision, and stereo-segmentation joint learning all reflect this geometry-aware trend \cite{fan2020sne,wang2021sneplus,gu2021cascaded,xue2021lidar,chang2022fast,fan2022learning,ma2023self,wu2024sgroadseg}. These methods better capture physical traversability and obstacle-free space, but still need stronger integration with temporal consistency, vehicle dynamics, traffic rules, and planning-level evaluation.

Overall, drivable space perception should be understood as functional road-space inference rather than generic semantic segmentation. Major challenges include distinguishing visually similar but non-accessible regions, handling dynamic obstacles and occlusions, modeling uncertainty in BEV or free-space maps, and evaluating whether the predicted region is actually useful for safe motion planning.

\begin{table*}[t]
\renewcommand\arraystretch{1.12}
\caption{\textbf{Representative studies on drivable space perception.}
The input icons indicate the input modalities used by each method:
\includegraphics[width=0.018\textwidth]{fig/icon/rgb.png} denotes camera input,
\includegraphics[width=0.018\textwidth]{fig/icon/lidar.png} denotes LiDAR input, and
\includegraphics[width=0.018\textwidth]{fig/icon/radar.png} denotes radar input.
}
\centering
\scriptsize
\setlength{\tabcolsep}{1.2mm}
\resizebox{\textwidth}{!}{
\begin{tabular}{p{2.3cm} p{1.1cm} C{1.4cm} p{2cm} p{2.6cm} p{3.9cm} p{6.5cm}}
\hline
\rowcolor{gray!10}
\textbf{Method} & \textbf{Venue} & \textbf{Input} & \textbf{Input Type} & \textbf{Output} & \textbf{Dataset} & \textbf{Key Idea} \\
\hline

\rowcolor{tpami_yellow!10}
\multicolumn{7}{l}{\textit{Image-based Drivable Area Segmentation}} \\
Teichmann et al. \cite{teichmann2018multinet} & IV'18 & \includegraphics[width=0.18\linewidth]{fig/icon/rgb.png} & Monocular & Road Segmentation & KITTI Road & Shares encoder across segmentation, detection, and classification. \\
Qian et al. \cite{qian2020dlt} & T-ITS'20 & \includegraphics[width=0.18\linewidth]{fig/icon/rgb.png} & Monocular & Drivable Segmentation & BDD100K & Jointly learns drivable areas with lanes and objects. \\
Wang et al. \cite{wang2021dynamic} & T-CYB'22 & \includegraphics[width=0.18\linewidth]{fig/icon/rgb.png} & RGB-D & Drivable Segmentation & GMRP, KITTI Road & Fuses RGB-D cues for drivable-area and anomaly detection. \\
Han et al. \cite{han2024epurate} & T-ITS'24 & \includegraphics[width=0.18\linewidth]{fig/icon/rgb.png} & Monocular & Road Segmentation & KITTI Road, Cityscapes, R2D & Refines uncertain boundaries with attention and spatial cues. \\
Sheng et al. \cite{sheng2024edgeunet} & T-ITS'25 & \includegraphics[width=0.18\linewidth]{fig/icon/rgb.png} & Monocular & Road/Lane Segmentation & BDD100K, TuSimple, KITTI & Uses edge guidance for drivable and lane segmentation. \\
Ma et al. \cite{ma2025task} & ICRA'25 & \includegraphics[width=0.18\linewidth]{fig/icon/rgb.png} & Monocular & Drivable Segmentation & KITTI Road & Selects SAM proposals using CLIP guidance. \\
Butt et al. \cite{butt2025r} & IJCV'25 & \includegraphics[width=0.18\linewidth]{fig/icon/rgb.png} & Monocular & Road Segmentation & R$^{2}$S100K & Introduces semi-supervised road-region benchmark. \\

\hline
\rowcolor{tpami_yellow!10}
\multicolumn{7}{l}{\textit{BEV and Occupancy-based Drivable Space Perception}} \\
Lu et al. \cite{lu2019monocular} & RA-L'19 & \includegraphics[width=0.18\linewidth]{fig/icon/rgb.png} & Monocular & BEV Occupancy & Cityscapes, KITTI & Predicts semantic occupancy from front-view images. \\
Mani et al. \cite{mani2020monolayout} & WACV'20 & \includegraphics[width=0.18\linewidth]{fig/icon/rgb.png} & Monocular & BEV Occupancy & KITTI, Argoverse & Hallucinates amodal BEV occupancy from one image. \\
Yang et al. \cite{yang2021projecting} & CVPR'21 & \includegraphics[width=0.18\linewidth]{fig/icon/rgb.png} & Monocular & BEV Occupancy & KITTI, Argoverse & Projects front-view features into BEV layout. \\
Popov et al. \cite{popov2023nvradarnet} & ICRA'23 & \includegraphics[width=0.18\linewidth]{fig/icon/radar.png} & Multi-radar & BEV Free-space & Private & Detects radar obstacles and BEV free space. \\
Liu et al. \cite{liu2024monocular} & TPAMI'24 & \includegraphics[width=0.18\linewidth]{fig/icon/rgb.png} & Monocular & BEV Occupancy & KITTI, Argoverse & Learns front-to-top projection for BEV perception. \\
Wen et al. \cite{wen2024upbev} & T-IV'24 & \includegraphics[width=0.18\linewidth]{fig/icon/lidar.png} & LiDAR & BEV Drivable Map & KITTI Road, SemanticKITTI & Uses polar BEV for LiDAR drivable maps. \\
Xiong et al. \cite{xiong2026bevcar} & CVPR'26 & \includegraphics[width=0.18\linewidth]{fig/icon/rgb.png} & Monocular & BEV Semantic Map & nuScenes, Argoverse & Context-aware rasterization for BEV semantic segmentation. \\

\hline
\rowcolor{tpami_yellow!10}
\multicolumn{7}{l}{\textit{3D and Multi-modal Free-space Perception}} \\
Fan et al. \cite{fan2020sne} & ECCV'20 & \includegraphics[width=0.18\linewidth]{fig/icon/rgb.png} & RGB-D & Free-space Segmentation & KITTI Road, SYNTHIA, R2D & Fuses RGB and normals for free-space segmentation. \\
Wang et al. \cite{wang2021sneplus} & IROS'21 & \includegraphics[width=0.18\linewidth]{fig/icon/rgb.png} & RGB-D & Free-space Segmentation & KITTI Road, R2D & Improves normal estimation and deep supervision. \\
Gu et al. \cite{gu2021cascaded} & ICRA'21 & \includegraphics[width=0.18\linewidth]{fig/icon/rgb.png}~\includegraphics[width=0.18\linewidth]{fig/icon/lidar.png} & Monocular/LiDAR & Road Segmentation & KITTI Road & Cascades LiDAR and camera cues for road segmentation. \\
Xue et al. \cite{xue2021lidar} & IROS'21 & \includegraphics[width=0.18\linewidth]{fig/icon/lidar.png} & LiDAR & Drivable Segmentation & SemanticKITTI, Private & Completes LiDAR drivable regions using spatial inference. \\
Chang et al. \cite{chang2022fast} & ICRA'22 & \includegraphics[width=0.18\linewidth]{fig/icon/rgb.png} & RGB-D & Road Segmentation & KITTI Road, Cityscapes & Uses uncertainty-aware RGB-D road fusion. \\
Fan et al. \cite{fan2022learning} & T-Mech'22 & \includegraphics[width=0.18\linewidth]{fig/icon/rgb.png} & Stereo & Free-space Segmentation & KITTI Road, SYNTHIA, SYN-Stereo & Augments stereo free-space learning with homographies. \\
Ma et al. \cite{ma2023self} & IROS'23 & \includegraphics[width=0.18\linewidth]{fig/icon/rgb.png}~\includegraphics[width=0.18\linewidth]{fig/icon/lidar.png} & Monocular/LiDAR & Drivable Segmentation & KITTI, KITTI-CARLA & Generates labels from LiDAR depth cues. \\
Wu et al. \cite{wu2024sgroadseg} & ICRA'24 & \includegraphics[width=0.18\linewidth]{fig/icon/rgb.png} & Stereo & Free-space Segmentation & KITTI Road & Shares stereo and segmentation encoders. \\
Chen et al. \cite{chen2025depth} & IROS'25 & \includegraphics[width=0.18\linewidth]{fig/icon/rgb.png} & RGB-D & Road Segmentation & KITTI Road, KITTI-360, Cityscapes & Models depth interactions for RGB-D road segmentation. \\
Chen et al. \cite{chen2026lidardhmt} & WACV'26 & \includegraphics[width=0.18\linewidth]{fig/icon/lidar.png} & LiDAR & Free-space Segmentation & KITTI Road & Adapts LiDAR masks for robust free-space segmentation. \\
\hline
\end{tabular}}
\vspace{-12pt}
\label{tab:drivable_space}
\end{table*}

\subsubsection{\textbf{Parking Slot Detection}}

Parking slot detection is a functional-spatial infrastructure perception task that aims to localize structured parking units available for low-speed maneuvering and automated parking. Unlike lane detection, which estimates longitudinal road guidance for normal driving, or drivable-space perception, which predicts general traversable regions, parking slot detection focuses on spatial units with explicit parking functionality. A parking slot is therefore more than a line pattern on the ground: it is jointly defined by slot markings, boundary segments, corner points, entrance lines, geometric layout, slot type, and, in many practical cases, occupancy or occlusion context. This functional interpretation makes parking slots a bridge between local infrastructure perception and downstream parking-space selection, local map construction, trajectory generation, and final vehicle alignment. Table~\ref{tab:parking_slot} summarizes representative studies from image-based and surround-view detection to structured and occupancy-aware parking-space perception.

Early learning-based methods mainly formulate the task in the image space, where parking slots are inferred from visual primitives such as marking points, line segments, entrance lines, or corners. DeepPS first establishes a DCNN-based parking slot detection pipeline and the PS2.0 benchmark, promoting the transition from handcrafted marking extraction to supervised visual slot perception \cite{zhang2018vision}. Subsequent methods improve robustness by directly regressing directional marking points, slot vertices, T-shaped center representations, or fixed anchor points, thereby reducing the dependence on brittle line-detection heuristics \cite{huang2019dmpr,wu2020psdet,bui2023transformer,zheng2024tpsd,jeong2025labcl}. These methods are effective when slot markings are visible, but their image-plane representation may still be sensitive to perspective distortion, partial observation, and inconsistent slot geometry.

Another line of work exploits surround-view cameras, fisheye cameras, inverse perspective mapping, and BEV representations. Since parking maneuvers are naturally planned in a vehicle-centric top-down coordinate frame, surround-view and BEV formulations provide more maneuver-friendly slot localization than single-view appearance detection. Surround-view methods detect parking slots from stitched or projected near-field imagery, while fisheye-based methods address distortion and large field-of-view coverage in raw camera inputs \cite{zinelli2019deep,fu2024parkscape}. Recent BEV and polygonal formulations further move beyond marking-level detection by directly predicting top-view slot geometry or polygonal slot boundaries, improving perception-to-planning consistency \cite{duan2025bevpolynet,wang2023holistic}.

The most structured formulations treat parking slots as geometric, relational, and occupancy-aware spatial entities rather than isolated visual primitives. Occupancy-aware detectors jointly estimate slot location and availability, while graph-based and component-linkage methods model relations among marking points or slot components to infer valid slot structures without heavy handcrafted post-processing \cite{li2020vacant,min2021attentional,suhr2021end,bui2023one}. Recent extensions further emphasize zero-shot mask extraction, semi-supervised learning, and real-world robustness under occlusion, damaged markings, dense layouts, or illumination changes \cite{zhai2024sam,zhang2025advancing}. In this sense, parking slot detection becomes a local infrastructure understanding problem: the system must reason about geometry, accessibility, and functional usability simultaneously.

Despite high benchmark scores on controlled datasets, parking slot perception remains difficult in real environments due to worn or missing markings, slanted and irregular slots, fisheye distortion, occlusion by vehicles or pedestrians, cross-site generalization, and the lack of unified representations that connect detection outputs directly with low-speed planning and local parking maps.

\begin{table*}[t]
\renewcommand\arraystretch{1.12}
\caption{\textbf{Representative studies on parking slot detection.}
The input icon indicates the input modality used by each method:
\includegraphics[width=0.018\textwidth]{fig/icon/rgb.png} denotes camera input.
}
\centering
\scriptsize
\setlength{\tabcolsep}{1.2mm}
\resizebox{\textwidth}{!}{
\begin{tabular}{p{2.4cm} p{1.8cm} C{1.4cm} p{2.2cm} p{2.7cm} p{1.8cm} p{5.5cm}}
\hline
\rowcolor{gray!10}
\textbf{Method} & \textbf{Venue} & \textbf{Input} & \textbf{Input Type} & \textbf{Output} & \textbf{Dataset} & \textbf{Key Idea} \\
\hline

\rowcolor{tpami_yellow!10}
\multicolumn{7}{l}{\textit{Image-based and Surround-view Parking Slot Detection}} \\
Zhang et al. \cite{zhang2018vision} & TIP'18 & \includegraphics[width=0.18\linewidth]{fig/icon/rgb.png} & Surround Monocular & Slot Instances & PS2.0 & Infers slots from DCNN marking points. \\
Huang et al. \cite{huang2019dmpr} & ICME'19 & \includegraphics[width=0.18\linewidth]{fig/icon/rgb.png} & Surround Monocular & Slot Instances & PS2.0 & Regresses directional marking points for slot pairing. \\
Zinelli et al. \cite{zinelli2019deep} & IV'19 & \includegraphics[width=0.18\linewidth]{fig/icon/rgb.png} & Surround Fisheye & Slot Instances & private & Detects slots from surround-view images. \\
Wu et al. \cite{wu2020psdet} & IV'20 & \includegraphics[width=0.18\linewidth]{fig/icon/rgb.png} & Surround Monocular & Slot Instances & PSDD, PS2.0 & Regresses vertices with circular descriptors. \\
Bui et al. \cite{bui2023transformer} & IEEE Access'23 & \includegraphics[width=0.18\linewidth]{fig/icon/rgb.png} & Surround Monocular & Slot Instances & PS2.0, SNU & Applies DETR with fixed anchor points. \\
Zheng et al. \cite{zheng2024tpsd} & JRTIP'24 & \includegraphics[width=0.18\linewidth]{fig/icon/rgb.png} & Surround Monocular & Slot Instances & PS2.0 & Uses T-shape keypoints for arbitrary-angle slots. \\
Fu et al. \cite{fu2024parkscape} & TIM'24 & \includegraphics[width=0.18\linewidth]{fig/icon/rgb.png} & Surround Fisheye & Slot Keypoints & ParkScape & Benchmarks fisheye slots with distortion-aware fusion. \\
Jeong et al. \cite{jeong2025labcl} & ICRA'25 & \includegraphics[width=0.18\linewidth]{fig/icon/rgb.png} & Surround Monocular & Slot Instances & PS2.0 & Balances junction learning with contrastive prototypes. \\
Duan et al. \cite{duan2025bevpolynet} & RA-L'25 & \includegraphics[width=0.18\linewidth]{fig/icon/rgb.png} & Surround Monocular & Slot Polygons & PS2.0 & Predicts BEV slot polygons end-to-end. \\
Zhang et al. \cite{zhang2026synthetic} & JICV'26 & \includegraphics[width=0.18\linewidth]{fig/icon/rgb.png} & Surround Monocular & Slot Instances & PSV, Boden AVM & Selects synthetic data for AVM slot detection. \\

\hline
\rowcolor{tpami_yellow!10}
\multicolumn{7}{l}{\textit{Structured and Occupancy-aware Parking Space Perception}} \\
Li et al. \cite{li2020vacant} & Sensors'20 & \includegraphics[width=0.18\linewidth]{fig/icon/rgb.png} & Surround Monocular & Slot Instances/Occupancy & PS2.0, PSV & Combines detection with occupancy classification. \\
Min et al. \cite{min2021attentional} & RA-L'21 & \includegraphics[width=0.18\linewidth]{fig/icon/rgb.png} & Surround Monocular & Slot Instances & PS2.0, PSV & Links marking points with graph attention. \\
Suhr et al. \cite{suhr2021end} & T-ITS'22 & \includegraphics[width=0.18\linewidth]{fig/icon/rgb.png} & Surround Monocular & Slot Instances/Occupancy & PS2.0 & Integrates global and local slot cues. \\
Bui et al. \cite{bui2023cnn} & IEEE Access'23 & \includegraphics[width=0.18\linewidth]{fig/icon/rgb.png} & Surround Monocular & Slot Instances/Occupancy & PS2.0, SNU & Uses region-specific multi-scale two-stage detection. \\
Bui et al. \cite{bui2023one} & ITS Magazine'23 & \includegraphics[width=0.18\linewidth]{fig/icon/rgb.png} & Surround Monocular & Slot Instances/Occupancy & PS2.0, SNU & Assembles slot components through learned linkages. \\
Wang et al. \cite{wang2023holistic} & IROS'23 & \includegraphics[width=0.18\linewidth]{fig/icon/rgb.png} & Surround Monocular & Slot Polygons & PS2.0 & Regresses top-view polygons in one step. \\
Zhai et al. \cite{zhai2024sam} & IV'24 & \includegraphics[width=0.18\linewidth]{fig/icon/rgb.png} & Surround Monocular & Slot Instances/Masks & PS2.0 & Uses SAM for zero-shot slot masks. \\
Zhang et al. \cite{zhang2025advancing} & T-ITS'25 & \includegraphics[width=0.18\linewidth]{fig/icon/rgb.png} & Surround Monocular & Slot Instances/Masks & CRPS-D, PS2.0 & Adds CRPS-D and semi-supervised baseline. \\

\hline
\end{tabular}}
\vspace{-12pt}
\label{tab:parking_slot}
\end{table*}

\subsection{Regulatory-Semantic Infrastructure Tasks}

\subsubsection{\textbf{Traffic Sign Recognition}}

Traffic sign recognition is a regulatory-semantic infrastructure perception task that aims to localize, recognize, and interpret vertical traffic signs as behavior-relevant driving constraints. Different from ordinary object detection, traffic signs are rule-bearing visual symbols: they encode speed limits, prohibitions, warnings, permissions, route guidance, and maneuver constraints, which affect speed control, behavior planning, map updating, and rule-compliant navigation \cite{liu2025exemplar}. Therefore, the output of traffic sign recognition is not merely an object category, but a compact semantic interface between visual perception and driving rules. This distinguishes it from traffic light perception, which focuses on dynamic signal states, road marking recognition, which parses surface-level regulatory markings, and lane detection, which estimates geometric lane structures. Table~\ref{tab:traffic_sign} summarizes representative studies from detection-recognition pipelines to semantic-aware and robust traffic sign understanding.

Early deep-learning-based traffic sign recognition mainly follows detection and recognition pipelines, the task is formulated as detecting sign instances and assigning closed-set class labels. Representative works move from real-time detection-classification systems to large-scale in-the-wild benchmarks and end-to-end detection-recognition frameworks \cite{yang2015towards,zhu2016traffic,lee2018simultaneous,tian2019traffic,yuan2019vssa,kamal2019automatic}. Large-scale datasets such as TT100K, MTSD, and DFG-TSD further shift the task from controlled cropped-sign classification toward real-world instance-level perception with small objects, cluttered backgrounds, illumination changes, and diverse sign taxonomies \cite{zhu2016traffic,tabernik2019deep,ertler2020mapillary}. In this category, traffic signs are commonly represented as bounding boxes, sign instances, masks, and class labels. Although such pipelines provide the basic perception interface for autonomous vehicles, they remain limited when sign categories are visually similar, geographically inconsistent, long-tailed, partially occluded, or degraded by weather and imaging conditions.

Recent research emphasizes semantic-aware and robust traffic sign recognition. This direction treats traffic signs as fine-grained symbolic entities whose visual appearance, textual content, pictograms, and traffic-rule meaning should be jointly modeled. Robustness-oriented studies investigate recognition under adverse illumination, blur, rain, fog, low resolution, occlusion, and other real-world degradations, using benchmarks such as CURE-TSR and CURE-TSD \cite{temel2017curetsr,temel2019traffic}. Long-tail and few-shot studies address rare categories and insufficient annotations through hierarchical label modeling, reweighting, and meta-learning \cite{ren2022metayolo,gao2022long,wang2024balanced}. Beyond closed-set label prediction, semantic-aware methods further parse sign components, textual cues, scene structure, and rule-level descriptions, enabling traffic sign understanding beyond simple classification \cite{zhu2017cascaded,guo2021learning,min2022traffic,guo2024signparser}. Recent CLIP- and VLM-based approaches introduce text prompts, semantic embeddings, and cross-region language-visual alignment to improve generalization across countries and sign standards \cite{zhao2025tsclip,gan2025crossdomain,zhao2026traffic}.

Despite substantial progress, traffic sign recognition still faces several open challenges, including tiny sign localization, taxonomy inconsistency across regions, rare and ambiguous categories, text-rich or compound signs, degradation under adverse environments, and the lack of unified evaluation protocols that connect visual recognition with rule-level driving semantics. From the road infrastructure perspective, future traffic sign perception should move beyond closed-set visual recognition toward sign-to-lane, sign-to-road-segment, and sign-to-rule grounding.

\begin{table*}[t]
\renewcommand\arraystretch{1.12}
\caption{\textbf{Representative studies on traffic sign recognition.}
The input icons indicate the input modalities used by each method:
\includegraphics[width=0.018\textwidth]{fig/icon/rgb.png} denotes camera input, and
\includegraphics[width=0.018\textwidth]{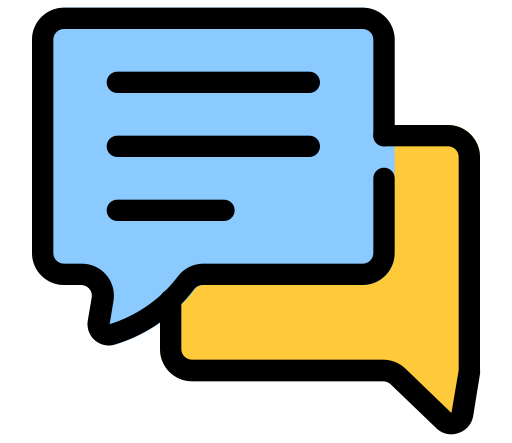} denotes text input.
}
\centering
\scriptsize
\setlength{\tabcolsep}{1.2mm}
\resizebox{\textwidth}{!}{
\begin{tabular}{p{2.0cm} p{1.4cm} C{1.4cm} p{1.6cm} p{2.6cm} p{2.9cm} p{5.6cm}}
\hline
\rowcolor{gray!10}
\textbf{Method} & \textbf{Venue} & \textbf{Input} & \textbf{Input Type} & \textbf{Output} & \textbf{Dataset} & \textbf{Key Idea} \\
\hline

\rowcolor{tpami_yellow!10}
\multicolumn{7}{l}{\textit{Traffic Sign Detection and Recognition}} \\
% Yang et al. \cite{yang2015towards} & T-ITS'16 & \includegraphics[width=0.18\linewidth]{fig/icon/rgb.png} & Monocular & Sign Boxes + Classes & GTSDB, Private & Color proposals with HOG classification. \\
Zhu et al. \cite{zhu2016traffic} & CVPR'16 & \includegraphics[width=0.18\linewidth]{fig/icon/rgb.png} & Monocular & Sign Boxes + Classes & TT100K & TT100K benchmark with CNN detection-classification. \\
Lee et al. \cite{lee2018simultaneous} & T-ITS'18 & \includegraphics[width=0.18\linewidth]{fig/icon/rgb.png} & Monocular & Sign Boxes + Geometry & GTSDB & Joint detection, pose, shape, and boundary estimation. \\
Zhu et al. \cite{zhu2017cascaded} & T-ITS'18 & \includegraphics[width=0.18\linewidth]{fig/icon/rgb.png} & Monocular & Text Sign Boxes & Private & Cascaded text-region segmentation and sign detection. \\
Yuan et al. \cite{yuan2019vssa} & TIP'19 & \includegraphics[width=0.18\linewidth]{fig/icon/rgb.png} & Monocular & Sign Boxes + Classes & GTSDB, TT100K & Vertical sequence attention for small signs. \\
Tian et al. \cite{tian2019traffic} & T-ITS'19 & \includegraphics[width=0.18\linewidth]{fig/icon/rgb.png} & Monocular & Sign Boxes + Classes & GTSDB, TT100K & Multi-scale recurrent attention for scale variation. \\
Kamal et al. \cite{kamal2019automatic} & T-ITS'20 & \includegraphics[width=0.18\linewidth]{fig/icon/rgb.png} & Monocular & Sign Masks + Classes & CURE-TSD & SegU-Net with Tversky-loss detection. \\
Tabernik et al. \cite{tabernik2019deep} & T-ITS'20 & \includegraphics[width=0.18\linewidth]{fig/icon/rgb.png} & Monocular & Sign Boxes + Classes & DFG-TS & Mask R-CNN for traffic-sign inventory recognition. \\
Ertler et al. \cite{ertler2020mapillary} & ECCV'20 & \includegraphics[width=0.18\linewidth]{fig/icon/rgb.png} & Monocular & Sign Boxes + Classes & MTSD & Global MTSD benchmark and detection baselines. \\
% Liu et al. \cite{liu2021tsingnet} & Neurocomputing'21 & \includegraphics[width=0.18\linewidth]{fig/icon/rgb.png} & Monocular & Sign Boxes + Classes & TT100K, STSD, DFG-TS & Scale-aware context learning for small signs. \\
Yu et al. \cite{yu2022traffic} & T-ITS'22 & \includegraphics[width=0.18\linewidth]{fig/icon/rgb.png} & Monocular & Sign Boxes + Classes & GTSDB, TT100K & Multi-frame YOLO and VGG recognition fusion. \\
Lin et al. \cite{lin2025tsdetr} & ICRA'25 & \includegraphics[width=0.18\linewidth]{fig/icon/rgb.png} & Monocular & Sign Boxes + Classes & TT100K, CCTSDB & Augments positives and negatives for DETR detection. \\
Chen et al. \cite{chen2025yolots} & T-ITS'25 & \includegraphics[width=0.18\linewidth]{fig/icon/rgb.png} & Monocular & Sign Boxes + Classes & TT100K, CCTSDB, GTSDB & Optimizes receptive fields and anchor-free fusion. \\
Xi et al. \cite{xi2025road} & T-ITS'25 & \includegraphics[width=0.18\linewidth]{fig/icon/rgb.png} & Monocular & Sign Classes & TT100K, GTSRB & Uses Siamese encoding for robust sign recognition. \\
\hline

\rowcolor{tpami_yellow!10}
\multicolumn{7}{l}{\textit{Semantic-aware and Robust Traffic Sign Recognition}} \\
Temel et al. \cite{temel2017curetsr} & NeurIPSW'17 & \includegraphics[width=0.18\linewidth]{fig/icon/rgb.png} & Monocular & Sign Classes & CURE-TSR & Controlled robustness benchmark for TSR. \\
Yu et al. \cite{yu2022traffic} & T-ITS'22 & \includegraphics[width=0.18\linewidth]{fig/icon/rgb.png} & Monocular & Sign Boxes + Classes & GTSDB, TT100K & Multi-frame YOLO and VGG recognition fusion. \\
Guo et al. \cite{guo2021learning} & ACM MM'21 & \includegraphics[width=0.18\linewidth]{fig/icon/rgb.png} & Monocular & Sign Semantics & CTSU & Component-relation parsing for sign understanding. \\
Ren et al. \cite{ren2022metayolo} & Appl. Sci.'22 & \includegraphics[width=0.18\linewidth]{fig/icon/rgb.png} & Monocular & Sign Boxes + Classes & TT100K & Meta-learning for few-shot sign detection. \\
Gao et al. \cite{gao2022long} & T-ITS'22 & \includegraphics[width=0.18\linewidth]{fig/icon/rgb.png} & Monocular & Sign Boxes + Classes & TT100K & Long-tailed detection with attentive fusion. \\
Min et al. \cite{min2022traffic} & T-ITS'22 & \includegraphics[width=0.18\linewidth]{fig/icon/rgb.png} & Monocular & Sign Classes & Private & Scene-structure-aware traffic sign recognition. \\
Guo et al. \cite{guo2024signparser} & IJCV'24 & \includegraphics[width=0.18\linewidth]{fig/icon/rgb.png} & Monocular & Sign Semantics & CTSU & End-to-end semantic sign parsing. \\
Wang et al. \cite{wang2024balanced} & T-ITS'24 & \includegraphics[width=0.18\linewidth]{fig/icon/rgb.png} & Monocular & Sign Boxes + Classes & TT100K, GTSDB & Balanced reweighting for long-tailed detection. \\
Zhao et al. \cite{zhao2025tsclip} & ICRA'25 & \includegraphics[width=0.18\linewidth]{fig/icon/rgb.png}~\includegraphics[width=0.18\linewidth]{fig/icon/text.png} & Monocular/Text & Sign Classes & CRTS & CLIP fine-tuning with traffic-rule prompts. \\
Gan et al. \cite{gan2025crossdomain} & KBS'25 & \includegraphics[width=0.18\linewidth]{fig/icon/rgb.png}~\includegraphics[width=0.18\linewidth]{fig/icon/text.png} & Monocular/Text & Sign Classes & Private & Multi-step VLM reasoning for zero-shot TSR. \\
Zhao et al. \cite{zhao2026traffic} & arXiv'26 & \includegraphics[width=0.18\linewidth]{fig/icon/rgb.png}~\includegraphics[width=0.18\linewidth]{fig/icon/text.png} & Monocular/Text & Sign Classes + Semantics & TS-1M & Million-scale diagnostic TSR benchmark. \\
\hline
\end{tabular}}
% \vspace{-6pt}
\label{tab:traffic_sign}
\end{table*}

\subsubsection{\textbf{Traffic Light Detection and State Recognition}}

Traffic light detection and state recognition is a regulatory-semantic infrastructure perception task that localizes traffic signal devices and interprets their dynamic states for rule-compliant autonomous driving. Unlike generic visual objects, traffic lights are dynamic rule-bearing infrastructure elements whose color, arrow, flashing, and pictogram states encode stop/go permissions, cautionary constraints, turning permissions, and movement-specific right-of-way rules. Therefore, this task cannot be reduced to small-object detection or simple color classification. It requires the joint perception of signal location, signal-head structure, active state, temporal consistency, and driving relevance at signalized intersections \cite{Jensen2016Vision,Behrendt2017Deep,Fregin2018DriveU}. Table~\ref{tab:traffic_light} summarizes representative studies from traffic-light detection and state recognition to relevance estimation.

Traffic light perception contains two closely coupled levels. The first is traffic light detection and localization, where traffic lights are represented as bounding boxes, signal instances, tracks, or signal-head regions. Early systems formulated traffic light perception as a camera-based detection--tracking--classification problem, while large-scale datasets such as Bosch and DTLD helped establish standardized detection and state annotations under real driving conditions \cite{Behrendt2017Deep,Fregin2018DriveU}. Recent works move beyond generic detection by emphasizing traffic-light-specific small-object localization, salient or ego-relevant signal detection, lightweight real-time inference, embedded deployment, sensor-failure robustness, multi-field-of-view perception, and adversarial robustness \cite{Greer2023Robust,Yao2023TLDetector,Polley2024TLDREADY,Li2024DynamicTLD,Huang2025DTLDetector,Lazzaroni2025Embedded,Polley2025ATLAS,Pavlitska2025Fool}. This evolution reflects a shift from merely finding traffic-light boxes to detecting signal elements that can support downstream regulatory interpretation in real vehicles.

The second level is traffic light state recognition, which extends localized signal perception into state-aware regulatory understanding. Its output representation includes red, yellow, green, off/unknown, arrow, flashing, pictogram, temporal, and relevance-aware states. Recent studies increasingly model individual signal bulbs, color and illumination variation, coarse-to-fine state hierarchies, video-based temporal cues, embedded real-time inference, nighttime enhancement, flashing-state dynamics, and ego-relevant signal selection \cite{Wu2023aUToLights,Chen2024Ensemble,Lin2024TwoStage,Yao2024Coarse,Fan2025Rockchip,Yao2025NightTime,Khaled2025FlashLightNet,Trinci2025Color,Belotti2025GraphMatching}. Compared with static traffic sign recognition and road marking recognition, traffic light perception is temporally dynamic: the same physical infrastructure element can impose different behavioral constraints over time. This makes temporal stability, state-transition filtering, arrow recognition, and relevance estimation essential for safe intersection behavior.

Open challenges include long-range small-object perception, adverse illumination, rare arrow and flashing states, cross-region signal design variation, false-positive suppression, temporal consistency, adversarial robustness, and ego-relevant signal selection. Although traffic light perception interacts with stop lines, crosswalks, lane topology, HD maps, and planning, its core role in road infrastructure perception remains dynamic regulatory-semantic interpretation rather than generic object detection, lane association, or intersection planning.

\begin{table*}[t]
\renewcommand\arraystretch{1.12}
\caption{\textbf{Representative studies on traffic light detection and state recognition.}
The input icons indicate the input modalities used by each method:
\includegraphics[width=0.018\textwidth]{fig/icon/rgb.png} denotes camera input, and
\includegraphics[width=0.018\textwidth]{fig/icon/lidar.png} denotes LiDAR input.
}
\centering
\scriptsize
\setlength{\tabcolsep}{1.2mm}
\resizebox{\textwidth}{!}{
\begin{tabular}{p{2.5cm} p{1.9cm} C{1.4cm} p{1.8cm} p{2.5cm} p{2.4cm} p{5.4cm}}
\hline
\rowcolor{gray!10}
\textbf{Method} & \textbf{Venue} & \textbf{Input} & \textbf{Input Type} & \textbf{Output} & \textbf{Dataset} & \textbf{Key Idea} \\
\hline

\rowcolor{tpami_yellow!10}
\multicolumn{7}{l}{\textit{Traffic Light Detection and Localization}} \\
Behrendt et al. \cite{Behrendt2017Deep} & ICRA'17 & \includegraphics[width=0.18\linewidth]{fig/icon/rgb.png} & Stereo & Boxes/States & BSTLD & Detects, tracks, and classifies traffic lights. \\
Fregin et al. \cite{Fregin2018DriveU} & ICRA'18 & \includegraphics[width=0.18\linewidth]{fig/icon/rgb.png} & Stereo & Boxes/States & DTLD & Publishes stereo benchmark with rich attributes. \\
Greer et al. \cite{Greer2023Robust} & IV'23 & \includegraphics[width=0.18\linewidth]{fig/icon/rgb.png} & Monocular & Boxes/Relevance & LAVA & Prioritizes ego-relevant traffic lights. \\
Yao et al. \cite{Yao2023TLDetector} & T-ITS'23 & \includegraphics[width=0.18\linewidth]{fig/icon/rgb.png} & Monocular & Boxes & TL2022, LaRA & Builds lightweight real-time detector. \\
Polley et al. \cite{Polley2024TLDREADY} & ITSC'24 & \includegraphics[width=0.18\linewidth]{fig/icon/rgb.png} & Monocular & Boxes/Relevance & BSTLD, LISA, DTLD & Estimates relevance from road arrows. \\
Li et al. \cite{Li2024DynamicTLD} & IEEE Sensors J.'24 & \includegraphics[width=0.18\linewidth]{fig/icon/rgb.png} & Monocular & Boxes & BSTLD, LISA, DTLD & Detects small lights with dynamic strategies. \\
Huang et al. \cite{Huang2025DTLDetector} & T-ITS'25 & \includegraphics[width=0.18\linewidth]{fig/icon/rgb.png} & Monocular & Boxes & YCTL2024, BSTLD & Uses lightweight detector with weather augmentation. \\
Lazzaroni et al. \cite{Lazzaroni2025Embedded} & IEEE Sensors J.'25 & \includegraphics[width=0.18\linewidth]{fig/icon/rgb.png} & Monocular & Boxes & Private & Adds failure mitigation on embedded platform. \\
Polley et al. \cite{Polley2025ATLAS} & IV'25 & \includegraphics[width=0.18\linewidth]{fig/icon/rgb.png} & Multi-Monocular & Boxes/States & ATLAS & Introduces multi-FOV traffic-light benchmark. \\
Pavlitska et al. \cite{Pavlitska2025Fool} & IV'25 & \includegraphics[width=0.18\linewidth]{fig/icon/rgb.png} & Monocular & Boxes & ATLAS, Private & Attacks detectors with printed patches. \\
Fan et al. \cite{Fan2025Rockchip} & IV'25 & \includegraphics[width=0.18\linewidth]{fig/icon/rgb.png} & Monocular & Boxes/States & Private & Uses video model on Rockchip hardware. \\
\hline

\rowcolor{tpami_yellow!10}
\multicolumn{7}{l}{\textit{Traffic Light State Recognition}} \\
Wu et al. \cite{Wu2023aUToLights} & CRV'23 & \includegraphics[width=0.18\linewidth]{fig/icon/rgb.png} & Multi-Monocular & Boxes/States & aUToLights & Filters state transitions with HMM. \\
Chen et al. \cite{Chen2024Ensemble} & IV'24 & \includegraphics[width=0.18\linewidth]{fig/icon/rgb.png} & Monocular & Bulb Boxes/States & Private & Combines color augmentation and ensembles. \\
% Lin et al. \cite{Lin2024TwoStage} & IEEE Access'24 & \includegraphics[width=0.18\linewidth]{fig/icon/rgb.png} & Monocular & Bulb Boxes/States & Private & Recognizes individual signal bulbs. \\
Yao et al. \cite{Yao2024Coarse} & T-ITS'24 & \includegraphics[width=0.18\linewidth]{fig/icon/rgb.png} & Monocular & Boxes/States & Private & Uses coarse-to-fine state recognition. \\
Yao et al. \cite{Yao2025NightTime} & T-SMC:S'25 & \includegraphics[width=0.18\linewidth]{fig/icon/rgb.png} & Monocular & Boxes/States & Night-TL, LISA & Guides detection with image enhancement. \\
Khaled et al. \cite{Khaled2025FlashLightNet} & Sensors'25 & \includegraphics[width=0.18\linewidth]{fig/icon/rgb.png} & Monocular & States & Private & Recognizes static and flashing states. \\
D{\'i}az-Acosta et al. \cite{DiazAcosta2026Tram} & T-ITS'26 & \includegraphics[width=0.18\linewidth]{fig/icon/rgb.png} & Monocular & Boxes/Light States & Private & Deploys tram perception on Jetson. \\
\hline

\rowcolor{tpami_yellow!10}
\multicolumn{7}{l}{\textit{Traffic Light Relevance Estimation}} \\
Belotti et al. \cite{Belotti2025GraphMatching} & ITSC'25 & \includegraphics[width=0.18\linewidth]{fig/icon/rgb.png}~\includegraphics[width=0.18\linewidth]{fig/icon/lidar.png} & Monocular/LiDAR & Boxes/Relevance & Private & Matches lights to LiDAR scene graphs. \\
Trinci et al. \cite{Trinci2025Color} & T-ITS'26 & \includegraphics[width=0.18\linewidth]{fig/icon/rgb.png} & Monocular & Boxes/States/Relevance & VZC-TLD, DTLD & Predicts state and ego relevance. \\
\hline
\end{tabular}}
\vspace{-12pt}
\label{tab:traffic_light}
\end{table*}

\subsubsection{\textbf{Road Marking Recognition}}

Road marking recognition aims to detect, classify, localize, and interpret rule-bearing symbols embedded on the road surface, including stop lines, crosswalks, road arrows, road text, speed-limit markings, lane-use control markings, bus-lane markings, yield markings, and other regulatory surface signs. Different from lane detection, which primarily estimates lane-line geometry, road marking recognition focuses on ground-level infrastructure elements that directly encode traffic rules and behavior constraints. A stop line specifies the action location for stopping, crosswalk markings indicate pedestrian-priority regions, arrows and lane-use markings constrain admissible maneuvers, and road text or speed markings provide symbolic instructions for driving decisions. Therefore, road markings should not be treated merely as road-surface textures or generic segmentation labels, but as regulatory-semantic infrastructure that supports rule-compliant planning and interpretable autonomous driving \cite{mathibela2015reading,amayo2018semantic}. Table~\ref{tab:road_marking} summarizes representative studies from 2D image-based recognition to BEV/map-oriented and LiDAR-based marking extraction.

According to sensing modality and spatial representation, existing studies can be summarized into two main streams: 2D image-based road marking recognition and 3D LiDAR-based road marking recognition. The former usually processes front-view, inverse-perspective, BEV, street-view, UAV, or aerial images, and predicts masks, bounding boxes, polygons, category labels, or map-oriented surface symbols. Early image-space methods mainly relied on segmentation, geometric grouping, and primitive-level interpretation, while deep learning methods further improved robustness through multitask learning, perspective guidance, benchmark-oriented training, and category-aware recognition. Representative image-based works such as VPGNet, TRoM, and CeyMo promote the transition from simple marking extraction toward non-lane road marking detection and semantic classification \cite{lee2017vpgnet,liu2017benchmark,jayasinghe2022ceymo}. Map-oriented and aerial-image methods further aggregate surface markings into lane-level or georeferenced map elements, including crosswalks, stop lines, road arrows, and ground signs, which are useful for localization, map updating, and planning \cite{qin2021roadmap,jang2022lane,barbero2025qaicedrone}. More recent vision-language methods further connect recognized markings with driving prompts and rule-level interpretation, indicating a shift from visual recognition to regulatory-semantic understanding \cite{sheng2026clip}.

The second stream focuses on 3D LiDAR and mobile laser scanning (MLS) point clouds. Compared with camera images, MLS-based methods exploit metric geometry, reflectance intensity, point density, and spatial continuity, making them particularly suitable for HD map production, road inventory, and precise localization. Earlier methods extract and classify road markings from intensity patterns and point clusters, while recent deep models introduce U-Net-like segmentation, capsule networks, graph attention, oriented bounding boxes, and instance-level boundary optimization to handle worn markings, non-uniform reflectance, sparse points, and complex marking shapes \cite{soilan2017segmentation,wen2019deep,ma2021capsule,mi2021two,fang2022graph,du2024rdmknet,li2025oriented,li2025context}.

Despite progress, road marking recognition remains challenging due to worn paint, occlusion, nighttime degradation, weather sensitivity, regional rule variation, long-tailed marking categories, and sensor-specific domain gaps. Future work should better integrate road markings with lane topology, traffic signs, traffic lights, HD maps, and planning, so that surface markings can be interpreted as explicit regulatory constraints rather than isolated visual patterns.

\begin{table*}[t]
\renewcommand\arraystretch{1.12}
\caption{\textbf{Representative studies on road marking recognition.}
The input icons indicate the input modalities used by each method:
\includegraphics[width=0.018\textwidth]{fig/icon/rgb.png} denotes camera input,
\includegraphics[width=0.018\textwidth]{fig/icon/lidar.png} denotes LiDAR input, and
\includegraphics[width=0.018\textwidth]{fig/icon/text.png} denotes text input.
}
\centering
\scriptsize
\setlength{\tabcolsep}{1.2mm}
\resizebox{\textwidth}{!}{
\begin{tabular}{p{2.2cm} p{2.3cm} C{1.4cm} p{1.6cm} p{2.5cm} p{1.7cm} p{6.2cm}}
\hline
\rowcolor{gray!10}
\textbf{Method} & \textbf{Venue} & \textbf{Input} & \textbf{Input Type} & \textbf{Output} & \textbf{Dataset} & \textbf{Key Idea} \\
\hline

\rowcolor{tpami_yellow!10}
\multicolumn{7}{l}{\textit{2D Image-based and BEV / Map-oriented Road Marking Recognition}} \\
Lee et al. \cite{lee2017vpgnet} & ICCV'17 & \includegraphics[width=0.18\linewidth]{fig/icon/rgb.png} & Monocular & Marking Segmentation & VPGNet & Uses vanishing-point-guided multitask recognition. \\
Liu et al. \cite{liu2017benchmark} & ITSC'17 & \includegraphics[width=0.18\linewidth]{fig/icon/rgb.png} & Monocular & Marking Segmentation & TRoM & Introduces 19-class road-marking benchmark. \\
Amayo et al. \cite{amayo2018semantic} & ITSC'18 & \includegraphics[width=0.18\linewidth]{fig/icon/rgb.png} & Monocular & Marking Classes & Private & Classifies primitives in inverse-perspective view. \\
Qin et al. \cite{qin2021roadmap} & ICRA'21 & \includegraphics[width=0.18\linewidth]{fig/icon/rgb.png} & Monocular & Marking Map & Private & Builds compact semantic maps for localization. \\
Jang et al. \cite{jang2022lane} & IEEE-CAA JAS'22 & \includegraphics[width=0.18\linewidth]{fig/icon/rgb.png} & Monocular & Marking Map & SeRM & Builds lane-level maps from segmented markings. \\
Jayasinghe et al. \cite{jayasinghe2022ceymo} & WACV'22 & \includegraphics[width=0.18\linewidth]{fig/icon/rgb.png} & Monocular & Marking Instances & CeyMo & Benchmarks non-standard markings with multiple annotations. \\
Guan et al. \cite{guan2022road} & JAG'22 & \includegraphics[width=0.18\linewidth]{fig/icon/rgb.png} & Aerial-view & Marking Segmentation & Private & Extracts UAV markings with attentive capsule FPN. \\
Barbero et al. \cite{barbero2025qaicedrone} & Photogramm. Rec.'25 & \includegraphics[width=0.18\linewidth]{fig/icon/rgb.png} & Aerial-view & Marking Map & Private & Maps UAV markings through U-Net and QGIS. \\
Sheng et al. \cite{sheng2026clip} & JAG'26 & \includegraphics[width=0.18\linewidth]{fig/icon/rgb.png}~\includegraphics[width=0.18\linewidth]{fig/icon/text.png} & Monocular/Text & Symbol Semantics & Private & Interprets symbols using CLIP and LLM prompts. \\

\hline
\rowcolor{tpami_yellow!10}
\multicolumn{7}{l}{\textit{3D LiDAR / MLS-based Road Marking Extraction and Recognition}} \\
Soil{\'a}n et al. \cite{soilan2017segmentation} & ISPRS JPRS'17 & \includegraphics[width=0.18\linewidth]{fig/icon/lidar.png} & LiDAR & Marking Instances & Private & Segments and classifies MLS reflective markings. \\
Wen et al. \cite{wen2019deep} & ISPRS JPRS'19 & \includegraphics[width=0.18\linewidth]{fig/icon/lidar.png} & LiDAR & Marking Map & Private & Completes MLS markings with deep learning. \\
Ma et al. \cite{ma2021capsule} & T-ITS'21 & \includegraphics[width=0.18\linewidth]{fig/icon/lidar.png} & LiDAR & Marking Segmentation & Private & Preserves shape cues with capsule networks. \\
Mi et al. \cite{mi2021two} & ISPRS JPRS'21 & \includegraphics[width=0.18\linewidth]{fig/icon/lidar.png} & LiDAR & 3D Marking & Private & Models MLS markings via two-stage localization. \\
Fang et al. \cite{fang2022graph} & JAG'22 & \includegraphics[width=0.18\linewidth]{fig/icon/lidar.png} & LiDAR & Marking Classes & Private & Classifies MLS markings with graph attention. \\
Du et al. \cite{du2024rdmknet} & T-ITS'24 & \includegraphics[width=0.18\linewidth]{fig/icon/lidar.png} & LiDAR & Marking Segmentation & RdmkNet, Toronto-RDMK & Provides large-scale classification and segmentation datasets. \\
Li et al. \cite{li2025oriented} & JSTARS'25 & \includegraphics[width=0.18\linewidth]{fig/icon/lidar.png} & LiDAR & Marking Instances & Private & Detects oriented markings using attention. \\
Li et al. \cite{li2025context} & Int. J. Digital Earth'25 & \includegraphics[width=0.18\linewidth]{fig/icon/lidar.png} & LiDAR & Marking Segmentation & Private & Segments instances with context and boundary cues. \\
\hline
\end{tabular}}
\vspace{-12pt}
\label{tab:road_marking}
\end{table*}

\section{Datasets and Evaluation Protocols}
\label{sec:datasets_evaluation}

\subsection{Representative Datasets for Road Infrastructure Perception}
\label{subsec:representative_datasets}

Datasets for road infrastructure perception should be distinguished from general autonomous-driving datasets that merely contain road scenes. A dataset is particularly relevant to this survey when its annotations explicitly describe road-organizing, function-constraining, or rule-bearing infrastructure elements, including lanes, curbs, road boundaries, lane topology, drivable regions, parking slots, traffic signs, traffic lights, and road markings. Accordingly, this subsection prioritizes infrastructure-specific datasets, while general-purpose benchmarks are included only when they provide widely used infrastructure annotations.

Benchmark maturity is highly uneven across infrastructure categories. Lane detection and traffic sign recognition have relatively mature datasets, whereas curb detection, road-boundary reasoning, lane topology, parking-space usability, traffic-light relevance, and road-marking semantics remain less standardized. Moreover, most datasets annotate visible elements, while only a few encode the functional, topological, and regulatory relations that make infrastructure behaviorally meaningful. Table~\ref{tab:representative_datasets_infrastructure} summarizes representative datasets in terms of task category, input modality, annotation form, scale, and main limitation.

\subsubsection{\textbf{Geometric-structural Infrastructure}}
Geometric-structural datasets show a clear transition from image-space lane localization to metric 3D lane geometry, vectorized map elements, and topology-aware infrastructure graphs. Early lane benchmarks such as TuSimple~\cite{tusimple2017lane}, CULane~\cite{pan2018spatial}, and LLAMAS~\cite{behrendt2019unsupervised} mainly support 2D lane or lane-marker localization. TuSimple focuses on regular highway scenes, CULane covers diverse urban conditions, and LLAMAS connects lane-marker annotation with HD-map projection and LiDAR-based localization. These datasets remain useful for image-space lane perception, but do not directly evaluate metric lane geometry, topology, or regulatory associations.

For planning-compatible lane perception, OpenLane~\cite{chen2022persformer} and ONCE-3DLanes~\cite{yan2022once} shift the benchmark target from visual lane localization to metric 3D lane geometry. They provide 3D lane annotations in vehicle or camera coordinates and cover diverse scenarios such as highways, bridges, tunnels, suburban roads, and downtown scenes. However, they still focus on geometric lane localization rather than lane connectivity or rule-level associations. By contrast, OpenLane-V2~\cite{wang2023openlanev2} and UrbanLaneGraph~\cite{buchner2023learning} move closer to infrastructure-level reasoning by introducing topology-oriented annotations, including lane-lane topology, lane-traffic-element topology, lane graphs, and map-compatible structures.

Road boundary and curb datasets are fewer but central to the proposed taxonomy because they target physical roadway limits rather than broad semantic classes. Oxford Road Boundaries~\cite{suleymanov2021oxford} provides visible and occluded road-boundary masks from Oxford RobotCar traversals, while Topo-boundary~\cite{xu2021topo} supports graph-oriented road-boundary detection from aerial imagery. For LiDAR-based curb detection, Uncertainty~\cite{jung2021uncertainty} provides public 32-line LiDAR data with BEV line- and point-style curb labels, and NRS~\cite{gao2023lcdet} extends public curb benchmarking to 128-line LiDAR point clouds in urban scenes. These datasets mainly support binary curb evaluation and lack full 3D semantic labeling. 3D-Curb~\cite{zhao2025curbnet} further provides SemanticKITTI-style point-cloud curb annotations and broader scene categories, partly addressing the lack of public LiDAR datasets with 3D curb labels. Together, these datasets show that boundary and curb perception remains less standardized than lane detection.

\subsubsection{\textbf{Functional-spatial Infrastructure}}
Functional-spatial datasets evaluate whether space is operationally usable for driving or parking. Compared with geometric-structural annotation, functional accessibility is harder to define because it depends on surfaces, obstacles, road condition, vehicle constraints, and maneuvering requirements. KITTI Road~\cite{fritsch2013new} remains a classical road-region benchmark, but its scale and diversity are limited. R2S100K~\cite{butt2025r} extends road-region segmentation to unstructured environments with potholes, water puddles, gravel, and irregular surfaces. ACDC~\cite{sakaridis2021acdc}, although not infrastructure-specific, is valuable for evaluating drivable-region perception under adverse weather and illumination. These datasets indicate that functional-spatial perception cannot be adequately evaluated on clean urban scenes alone.

Parking slot datasets are specialized functional-spatial benchmarks. DeepPS~\cite{zhang2018vision} provides surround-view parking images with marking-point and slot-instance annotations for automated parking. Context-based Parking Slot Detection~\cite{do2020context} emphasizes realistic parking geometry and contextual reasoning. ParkScape~\cite{fu2024parkscape} extends parking slot perception to fisheye cameras with corner-based labels, which is important for production surround-view systems. CRPS-D~\cite{zhang2025advancing} further addresses lighting changes, weather, damaged markings, and slanted slots. Compared with general drivable-area datasets, parking datasets better evaluate local structured-space perception and geometric usability, but remain weakly connected to occupancy reasoning and low-speed planning.

\subsubsection{\textbf{Regulatory-semantic Infrastructure}}
Regulatory-semantic datasets are rich in object-level labels but weak in rule-level grounding. Traffic sign datasets are the most established resources in this category. GTSRB~\cite{stallkamp2012man} and GTSDB~\cite{houben2013detection} provide early classification and detection benchmarks, but are geographically narrow and mainly focus on German traffic signs. TT100K~\cite{zhu2016traffic} supports in-the-wild traffic sign detection and classification for Chinese road scenes, while Mapillary Traffic Sign Dataset~\cite{ertler2020mapillary} provides global-scale fine-grained annotations. These datasets support regulatory-semantic recognition, but most still treat traffic signs as isolated visual objects rather than lane-associated or route-conditioned rules.

Traffic light datasets also focus on rule-bearing infrastructure, but differ in annotation richness and scale. LISA Traffic Light~\cite{philipsen2015learning} provides early annotations for traffic light detection. Bosch Small Traffic Lights~\cite{Behrendt2017Deep} emphasizes small traffic light detection and active-light state classification. DriveU Traffic Light Dataset~\cite{Fregin2018DriveU} provides boxes, states, attributes, and visibility labels for state-aware traffic-light understanding. However, most traffic-light datasets lack ego-lane relevance, lane-level signal association, and maneuver-level interpretation, which are critical at intersections.

Road marking datasets remain underdeveloped despite their importance for regulatory-semantic infrastructure. ApolloScape~\cite{huang2018apolloscape} provides lane-marking and road-scene annotations for map-compatible perception. CeyMo~\cite{jayasinghe2022ceymo} provides polygon, bounding-box, and segmentation-mask annotations for road-marking classes such as arrows and crosswalk-related markings. Nevertheless, most road-marking datasets annotate visual symbols rather than legal meanings, lane associations, or maneuver-level implications, so they evaluate recognition more than rule-aware understanding.

Overall, existing datasets reveal three structural gaps. First, benchmark maturity is uneven: lane and traffic sign datasets are relatively mature, while curb, road boundary, lane topology, parking slot, traffic-light relevance, and road-marking semantics remain less standardized. Second, most datasets evaluate isolated elements rather than their geometric, functional, and regulatory relationships. Third, future benchmarks should move beyond visible-element annotation toward integrated evaluation of geometry, functional space, regulatory semantics, topology, temporal consistency, and downstream planning relevance.

\begin{table*}[t]
\centering
\caption{\textbf{Representative infrastructure-specific datasets for road infrastructure perception.}
\Geo, \Func, and \Reg denote geometric-structural, functional-spatial, and regulatory-semantic infrastructure, respectively.
The input icons indicate the input modalities or priors provided by each dataset:
\includegraphics[width=0.018\textwidth]{fig/icon/rgb.png} denotes camera input,
\includegraphics[width=0.018\textwidth]{fig/icon/lidar.png} denotes LiDAR input,
\includegraphics[width=0.018\textwidth]{fig/icon/map.png} denotes HD map input,
\includegraphics[width=0.018\textwidth]{fig/icon/gnss.png} denotes GNSS input, and
\includegraphics[width=0.018\textwidth]{fig/icon/text.png} denotes text input.
}
\label{tab:representative_datasets_infrastructure}
\scriptsize
\renewcommand{\arraystretch}{1.1}
\setlength{\tabcolsep}{0.8mm}
\resizebox{\textwidth}{!}{
\begin{tabular}{p{2.2cm} p{1.4cm} p{2cm} C{1.5cm} p{3.1cm} p{2.4cm} p{1.5cm} p{3.9cm}}
\toprule
\rowcolor{gray!10}
\textbf{Dataset} 
& \textbf{Venue} 
& \textbf{Category / Task} 
& \textbf{Input} 
& \textbf{Input Type} 
& \textbf{Annotation} 
& \textbf{Scale} 
& \textbf{Main Limitation} \\
\midrule

TuSimple~\cite{tusimple2017lane} 
& Benchmark'17 
& \Geo~Lane Detection 
& \includegraphics[width=0.18\linewidth]{fig/icon/rgb.png}
& Monocular
& Points; Curves 
& 6.4K clips 
& Highway-biased; no topology labels. \\

CULane~\cite{pan2018spatial} 
& AAAI'18 
& \Geo~Lane Detection 
& \includegraphics[width=0.18\linewidth]{fig/icon/rgb.png}
& Monocular
& Curves; Masks 
& 133K frames 
& Image-space only; no metric topology. \\

LLAMAS~\cite{behrendt2019unsupervised} 
& ICCVW'19 
& \Geo~Lane Detection 
& \includegraphics[width=0.18\linewidth]{fig/icon/rgb.png}~\includegraphics[width=0.18\linewidth]{fig/icon/map.png}
& Monocular/HD Map
& Markers; Associations 
& 100K images 
& Projection errors; weak topology. \\

OpenLane~\cite{chen2022persformer} 
& ECCV'22 
& \Geo~Lane Detection 
& \includegraphics[width=0.18\linewidth]{fig/icon/rgb.png}
& Monocular
& 3D points 
& 200K frames 
& Limited relation labels. \\

ONCE-3DLanes~\cite{yan2022once} 
& CVPR'22 
& \Geo~Lane Detection 
& \includegraphics[width=0.18\linewidth]{fig/icon/rgb.png}~\includegraphics[width=0.18\linewidth]{fig/icon/lidar.png}
& Monocular/LiDAR
& 3D points 
& 211K frames 
& Geometry-focused; weak topology. \\

OpenLane-V2~\cite{wang2023openlanev2} 
& NeurIPS'23 
& \Geo~Lane Topology 
& \includegraphics[width=0.18\linewidth]{fig/icon/rgb.png}~\includegraphics[width=0.18\linewidth]{fig/icon/map.png}
& Surround-Monocular/HD Map
& Polylines; Topology 
& 2K scenes 
& Dataset-specific topology. \\

UrbanLaneGraph~\cite{buchner2023learning} 
& CVPR'23 
& \Geo~Lane Topology 
& \includegraphics[width=0.18\linewidth]{fig/icon/rgb.png}
& Aerial-view
& Nodes; Edges 
& 43K splits
& Offline aerial image setting. \\

Oxford Road~\cite{suleymanov2021oxford} 
& arXiv'21 
& \Geo~Road Boundary 
& \includegraphics[width=0.18\linewidth]{fig/icon/rgb.png}
& Stereo
& Boundary masks 
& 62.6K samples 
& Image-space only; lacks 3D curbs. \\

Topo-boundary~\cite{xu2021topo} 
& RA-L'21 
& \Geo~Road Boundary 
& \includegraphics[width=0.18\linewidth]{fig/icon/rgb.png}
& Aerial-view
& Masks; Graphs 
& 25.3K images 
& Offline aerial domain. \\

Uncertainty~\cite{jung2021uncertainty}
& ICRA'21
& \Geo~Curb Detection
& \includegraphics[width=0.18\linewidth]{fig/icon/lidar.png}
& LiDAR
& BEV line / point labels
& 5.2K frames
& No 3D curb labels; binary setting. \\

NRS~\cite{gao2023lcdet}
& IJCNN'23
& \Geo~Curb Detection
& \includegraphics[width=0.18\linewidth]{fig/icon/lidar.png}
& LiDAR
& BEV line labels
& 6.2K frames
& High-resolution LiDAR; no 3D labels. \\

3D-Curb~\cite{zhao2025curbnet} 
& T-ITS'25 
& \Geo~Curb Detection 
& \includegraphics[width=0.18\linewidth]{fig/icon/lidar.png}
& LiDAR
& 3D point labels 
& 7.1K frames 
& SemanticKITTI-based; limited regions. \\

\midrule

KITTI Road~\cite{fritsch2013new} 
& ITSC'13 
& \Func~Drivable Space 
& \includegraphics[width=0.18\linewidth]{fig/icon/rgb.png}
& Stereo
& Road masks 
& 579 images 
& Limited scale; few adverse cases. \\

Cityscapes~\cite{cordts2016cityscapes}
& CVPR'16
& \Func~Drivable Space
& \includegraphics[width=0.18\linewidth]{fig/icon/rgb.png}
& Stereo
& Semantic masks
& 25K images
& Urban-only; not infrastructure-specific. \\

BDD100K~\cite{yu2020bdd100k}
& CVPR'20
& \Func~Drivable Space
& \includegraphics[width=0.18\linewidth]{fig/icon/rgb.png}
& Monocular
& Drivable masks; Lanes
& 100K videos
& 2D labels; weak 3D grounding. \\

ACDC~\cite{sakaridis2021acdc} 
& ICCV'21 
& \Func~Drivable Space 
& \includegraphics[width=0.18\linewidth]{fig/icon/rgb.png}
& Monocular
& Semantic masks 
& 4K images 
& General semantic benchmark. \\

SemanticKITTI~\cite{behley2019semantickitti}
& ICCV'19
& \Func~Drivable Space
& \includegraphics[width=0.18\linewidth]{fig/icon/lidar.png}
& LiDAR
& Point labels
& 43K scans
& General semantics; no lane topology. \\

R$^{2}$S100K~\cite{butt2025r} 
& IJCV'25 
& \Func~Drivable Space 
& \includegraphics[width=0.18\linewidth]{fig/icon/rgb.png}
& Monocular
& Road masks 
& 100K images 
& Weak lane-level constraints. \\

PS2.0 / DeepPS~\cite{zhang2018vision} 
& TIP'18 
& \Func~Parking Slot 
& \includegraphics[width=0.18\linewidth]{fig/icon/rgb.png}
& Surround Monocular
& Points; Slots 
& 12.2K images 
& Limited environment diversity. \\

Context Parking~\cite{do2020context} 
& Access'20 
& \Func~Parking Slot 
& \includegraphics[width=0.18\linewidth]{fig/icon/rgb.png}
& Surround Monocular
& Boxes; Context 
& 22.8K images 
& Less standardized than PS2.0. \\

ParkScape~\cite{fu2024parkscape} 
& TIM'24 
& \Func~Parking Slot 
& \includegraphics[width=0.18\linewidth]{fig/icon/rgb.png}
& Surround Fisheye
& Corners; Slots 
& 10K images 
& Newer; limited adoption. \\

CRPS-D~\cite{zhang2025advancing} 
& T-ITS'25 
& \Func~Parking Slot 
& \includegraphics[width=0.18\linewidth]{fig/icon/rgb.png}
& Surround Monocular
& Labels; Masks 
& 35.4K images 
& New benchmark; adoption unverified. \\

\midrule

GTSRB~\cite{stallkamp2012man} 
& NN'12 
& \Reg~Traffic Sign 
& \includegraphics[width=0.18\linewidth]{fig/icon/rgb.png}
& Monocular
& Classes 
& 50K images 
& Country-specific; cropped setting. \\

GTSDB~\cite{houben2013detection} 
& IJCNN'13 
& \Reg~Traffic Sign 
& \includegraphics[width=0.18\linewidth]{fig/icon/rgb.png}
& Monocular
& Boxes; Classes 
& 900 images 
& Limited scale and diversity. \\

TT100K~\cite{zhu2016traffic} 
& CVPR'16 
& \Reg~Traffic Sign 
& \includegraphics[width=0.18\linewidth]{fig/icon/rgb.png}
& Monocular
& Boxes; Masks 
& 100K images 
& Region-specific; long-tailed. \\

Mapillary~\cite{ertler2020mapillary} 
& ECCV'20 
& \Reg~Traffic Sign 
& \includegraphics[width=0.18\linewidth]{fig/icon/rgb.png}
& Monocular
& Boxes; Classes 
& 100K images 
& No lane-level association. \\

TS-1M~\cite{zhao2026traffic}
& arXiv'26
& \Reg~Traffic Sign 
& \includegraphics[width=0.18\linewidth]{fig/icon/rgb.png}~\includegraphics[width=0.18\linewidth]{fig/icon/text.png}
& Monocular/Text
& Classes; Text semantics
& 1.2M images
& New benchmark; adoption unverified. \\

LISA~\cite{philipsen2015learning} 
& ITSC'15 
& \Reg~Traffic Light 
& \includegraphics[width=0.18\linewidth]{fig/icon/rgb.png}
& Stereo
& Boxes; States 
& 43K frames 
& Limited modern diversity. \\

Bosch Traffic~\cite{Behrendt2017Deep} 
& ICRA'17 
& \Reg~Traffic Light 
& \includegraphics[width=0.18\linewidth]{fig/icon/rgb.png}
& Monocular
& Boxes; States 
& 13.4K images 
& No ego-lane relevance. \\

DriveU Traffic~\cite{Fregin2018DriveU} 
& ICRA'18 
& \Reg~Traffic Light 
& \includegraphics[width=0.18\linewidth]{fig/icon/rgb.png}~\includegraphics[width=0.18\linewidth]{fig/icon/gnss.png}
& Stereo/GNSS
& Boxes; Attributes 
& 230K lights 
& Weak lane association. \\

LAVA~\cite{Greer2023Robust}
& IV'23
& \Reg~Traffic Light
& \includegraphics[width=0.18\linewidth]{fig/icon/rgb.png}
& Monocular
& Boxes; Relevance
& 11.4K images
& Relevance-focused; limited diversity. \\

ATLAS~\cite{Polley2025ATLAS}
& IV'25
& \Reg~Traffic Light
& \includegraphics[width=0.18\linewidth]{fig/icon/rgb.png}
& Multi-Monocular
& Boxes; States
& 33K images
& New dataset; German-city bias. \\

ApolloScape~\cite{huang2018apolloscape} 
& CVPRW'18 
& \Reg~Road Marking 
& \includegraphics[width=0.18\linewidth]{fig/icon/rgb.png}~\includegraphics[width=0.18\linewidth]{fig/icon/gnss.png}~\includegraphics[width=0.18\linewidth]{fig/icon/map.png}
& Monocular/GNSS/HD Map
& Markings; Masks 
& 140K images 
& Weak rule-level semantics. \\

VPGNet~\cite{lee2017vpgnet}
& ICCV'17
& \Reg~Road Marking
& \includegraphics[width=0.18\linewidth]{fig/icon/rgb.png}
& Monocular
& Markings; Lanes
& 20K images
& Focused on adverse weather scenes. \\

CeyMo~\cite{jayasinghe2022ceymo} 
& WACV'22 
& \Reg~Road Marking 
& \includegraphics[width=0.18\linewidth]{fig/icon/rgb.png}
& Monocular
& Polygons; Masks 
& 2.9K images 
& Limited classes; no lane grounding. \\

\bottomrule
\end{tabular}}
\vspace{-12pt}
\end{table*}

\subsection{Representation-aware Evaluation Protocols}
\label{subsec:evaluation_protocols_infrastructure}

Although road infrastructure perception spans diverse tasks, its evaluation protocols are more coherently understood through output representations and downstream utility than through task names alone. As summarized in Table~\ref{tab:evaluation_metrics_infrastructure}, different outputs, including labels, boxes, masks, keypoints, curves, polylines, BEV grids, vector maps, occupancy maps, and topology graphs, require distinct matching rules and error measures. The same metric family may apply across tasks when their outputs share a representation, while the same task may require different metrics as its representation changes from image-space masks to BEV grids, vectorized elements, or topology graphs.

Classification, detection, and segmentation metrics remain the dominant protocols for regulatory-semantic objects and dense infrastructure regions. Accuracy, macro-F1, balanced accuracy, and per-class recall are commonly used for traffic sign classification, traffic light state recognition, and road-marking recognition. Precision, recall, AP, and mAP are used for traffic signs, traffic lights, object-like parking slots, and road markings. Pixel accuracy, IoU, mIoU, Dice/F1, boundary F-score, and MaxF are used for drivable areas, road masks, lane masks, curb masks, crosswalks, stop lines, and semantic road parsing. These metrics measure local recognition, localization, or region overlap, but are insufficient for lane relevance, rule grounding, topology, temporal consistency, and planning compatibility.

Geometric infrastructure elements require structure-aware metrics. Lane detection, 3D lane detection, road-boundary detection, curb extraction, parking slot localization, stop-line detection, and vectorized road marking recognition use point-wise distance, endpoint error, lateral and longitudinal error, Chamfer distance, Hausdorff distance, or distance-threshold precision/recall/F1. These metrics are more suitable than pixel IoU for thin structures because they directly measure spatial deviation between predicted and ground-truth curves, points, or polylines. However, they remain limited in assessing semantic validity, topological connectivity, and rule consistency. A lane polyline may be close to the ground truth but still connect to the wrong branch at an intersection.

BEV, occupancy, and map-element metrics are increasingly important as infrastructure perception moves toward planning-compatible coordinate frames. BEV IoU, occupancy IoU, grid accuracy, vector-map AP, polyline AP, Chamfer-distance-based AP, and map-element matching are used for online HD map construction, BEV lane perception, drivable space estimation, crosswalk detection, road-boundary prediction, lane-divider extraction, and centerline estimation. These metrics assess whether infrastructure elements are correctly placed in a metric top-down frame, but are sensitive to calibration, ego-pose errors, grid resolution, coordinate-frame definitions, and dataset-specific annotation conventions. More importantly, high BEV or vector-map scores do not necessarily indicate downstream utility unless the predicted map improves localization, routing, planning, or rule compliance.

Topology and relation metrics evaluate relationships among infrastructure elements. Graph precision, graph recall, topology F1, adjacency accuracy, connectivity accuracy, lane-lane relation accuracy, and lane-traffic-element association accuracy are used for lane topology, intersection layout perception, lane graph construction, and lane-level traffic sign/light association. These metrics are essential because autonomous driving requires not only detecting lanes, signs, and signals, but also understanding lane connectivity, legal maneuvers, and ego-relevant regulatory elements. However, topology evaluation remains less standardized than detection or segmentation because datasets define nodes, edges, lane segments, connectors, and traffic-element associations differently.

Robustness and generalization metrics are necessary because road infrastructure is strongly domain-dependent. Lane marking styles, sign designs, signal layouts, curb structures, road textures, and marking maintenance vary across regions, weather, and regulation systems. Cross-domain performance drop, relative degradation, rare-class recall, long-tail F1, adverse-condition mIoU, and performance stratified by weather, lighting, region, or visibility are therefore important for deployability. A model performing well on a single benchmark may still fail under different regional rules, degraded markings, night scenes, adverse weather, construction zones, or sensor degradation.

Downstream-oriented evaluation remains the least mature but most important direction. Infrastructure perception ultimately supports localization, online map updating, prediction, motion planning, and rule-compliant control. Relevant criteria include planning compatibility, map update quality, route-level correctness, off-road rate, collision rate, lane compliance, rule compliance, temporal stability, and closed-loop driving performance. Although difficult to standardize because they require downstream modules, simulators, or real-world driving logs, such metrics are necessary because infrastructure errors have unequal behavioral consequences: a small road-mask error may be harmless, whereas an incorrect lane-traffic-light association or missing stop line may invalidate a maneuver.

\begin{table*}[t]
\centering
\caption{\textbf{Evaluation metric families for road infrastructure perception.}
The table summarizes representative metrics, formulas, symbol definitions, and their representation-level applicability.}
\label{tab:evaluation_metrics_infrastructure}
\scriptsize
\renewcommand{\arraystretch}{0.95}
\setlength{\tabcolsep}{0.55mm}
\resizebox{\textwidth}{!}{%
\begin{tabular}{
>{\raggedright\arraybackslash}p{1.45cm}
>{\raggedright\arraybackslash}p{2.35cm}
>{\raggedright\arraybackslash\footnotesize}p{6.45cm}
>{\raggedright\arraybackslash}p{3.8cm}
>{\raggedright\arraybackslash}p{4.3cm}
}

\toprule
\rowcolor{gray!10}
\multicolumn{2}{l}{\textbf{Metric Family}} 
& \textbf{Formula} 
& \textbf{Symbol Definition} 
& \textbf{Output / Limitation} \\
\midrule

\multirow{7}{=}{Classification}
& Accuracy
& $\mathrm{Acc}=\frac{1}{N}\sum_{i=1}^{N}\mathbb{I}(\hat{y}_i=y_i)$
& $N$: samples; $\hat{y}_i$: predicted label; $y_i$: ground-truth label
& Class labels for signs, lights, and markings; biased by frequent classes. \\
\cmidrule(lr){2-5}

& Precision / Recall / F1
& $\mathrm{Prec}=\frac{TP}{TP+FP}$, 
$\mathrm{Rec}=\frac{TP}{TP+FN}$, 
$\mathrm{F1}=\frac{2\mathrm{Prec}\cdot\mathrm{Rec}}{\mathrm{Prec}+\mathrm{Rec}}$
& $TP$: true positives; $FP$: false positives; $FN$: false negatives
& Class-level recognition; weak for localization and lane-level relevance. \\
\cmidrule(lr){2-5}

& Macro-F1 / Balanced Accuracy
& $\mathrm{MacroF1}=\frac{1}{C}\sum_{c=1}^{C}\mathrm{F1}_c$, 
$\mathrm{BAcc}=\frac{1}{C}\sum_{c=1}^{C}\mathrm{Rec}_c$
& $C$: classes; $\mathrm{F1}_c$: class-wise F1; $\mathrm{Rec}_c$: class-wise recall
& Long-tail signs and rare markings; sensitive to noisy rare-class labels. \\

\midrule

\multirow{4}{=}{Detection}
& AP / mAP
& $\mathrm{AP}_c=\int_{0}^{1}p_c(r)\,dr$, 
$\mathrm{mAP}=\frac{1}{C}\sum_{c=1}^{C}\mathrm{AP}_c$
& $p_c(r)$: precision-recall curve for class $c$; $C$: classes
& Boxes for signs, lights, slots, and markings; weak for rule relevance. \\
\cmidrule(lr){2-5}

& IoU Matching
& $\mathrm{IoU}(B,\hat{B})=\frac{|B\cap \hat{B}|}{|B\cup \hat{B}|}$
& $B$: ground-truth box or region; $\hat{B}$: predicted box or region
& Object-like regions; unsuitable for thin or elongated infrastructure. \\

\midrule

\multirow{6}{=}{Segmentation}
& IoU / mIoU
& $\mathrm{IoU}_c=\frac{TP_c}{TP_c+FP_c+FN_c}$, 
$\mathrm{mIoU}=\frac{1}{C}\sum_{c=1}^{C}\mathrm{IoU}_c$
& $TP_c,FP_c,FN_c$: pixel / point / grid counts for class $c$; $C$: classes
& Masks for roads, curbs, lanes, and markings; weak for boundary shifts. \\
\cmidrule(lr){2-5}

& Dice / Region F1
& $\mathrm{Dice}=\frac{2|P\cap G|}{|P|+|G|}$
& $P$: predicted region; $G$: ground-truth region
& Region masks; weak for topology and geometric continuity. \\
\cmidrule(lr){2-5}

& Boundary F-score
& $\mathrm{BF}=\frac{2P_bR_b}{P_b+R_b}$
& $P_b$: boundary precision; $R_b$: boundary recall
& Boundary masks for curbs, lanes, and crosswalks; tolerance-dependent. \\

\midrule

\multirow{8}{=}{Keypoint / Curve / Boundary}
& Chamfer Distance
& $d_{\mathrm{CD}}(P,G)=\frac{1}{|P|}\sum_{p\in P}\min_{g\in G}\|p-g\|_2+\frac{1}{|G|}\sum_{g\in G}\min_{p\in P}\|g-p\|_2$
& $P,G$: predicted and ground-truth point sets or polylines
& Points, curves, lanes, curbs, and boundaries; ignores connectivity. \\
\cmidrule(lr){2-5}

& Hausdorff Distance
& $d_{\mathrm{H}}(P,G)=\max\{\sup_{p\in P}\inf_{g\in G}\|p-g\|_2$, $\sup_{g\in G}\inf_{p\in P}\|g-p\|_2\}$
& $P,G$: predicted and ground-truth point sets
& Boundary and curve localization; sensitive to outliers and annotation noise. \\
\cmidrule(lr){2-5}

& Distance-threshold Precision / Recall
& $\mathrm{Prec}_{\delta}=\frac{|\{p\in P:\min_{g\in G}\|p-g\|_2<\delta\}|}{|P|}$, 
$\mathrm{Rec}_{\delta}=\frac{|\{g\in G:\min_{p\in P}\|g-p\|_2<\delta\}|}{|G|}$
& $\delta$: distance threshold; $P,G$: predicted and ground-truth samples
& Lanes, slot corners, and stop lines; threshold-sensitive and topology-agnostic. \\

\midrule

\multirow{5}{=}{BEV / Occupancy / Map}
& BEV IoU / Occupancy IoU
& $\mathrm{IoU}_{\mathrm{BEV}}=\frac{|\hat{M}\cap M|}{|\hat{M}\cup M|}$
& $\hat{M},M$: predicted and ground-truth BEV / occupancy maps
& BEV grids and occupancy maps; sensitive to calibration and resolution. \\
\cmidrule(lr){2-5}

& Vector-map AP / Polyline AP
& $\mathrm{AP}_{\mathrm{map}}=\int_{0}^{1}p_{\mathrm{match}}(r)\,dr$
& $p_{\mathrm{match}}(r)$: precision-recall curve after map-element matching
& Vectorized lanes, boundaries, and crosswalks; dataset-specific conventions. \\

\midrule

\multirow{4}{=}{Topology / Relation}
& Topology F1
& $\mathrm{F1}_{\mathrm{topo}}=\frac{2\mathrm{Prec}_{E}\mathrm{Rec}_{E}}{\mathrm{Prec}_{E}+\mathrm{Rec}_{E}}$
& $\mathrm{Prec}_{E}$: edge precision; $\mathrm{Rec}_{E}$: edge recall
& Lane graphs and intersection topology; graph matching is weakly standardized. \\
\cmidrule(lr){2-5}

& Adjacency / Connectivity Accuracy
& $\mathrm{Acc}_{\mathrm{rel}}=\frac{1}{|\mathcal{R}|}\sum_{r\in\mathcal{R}}\mathbb{I}(\hat{r}=r)$
& $\mathcal{R}$: ground-truth relation set; $\hat{r}$: predicted relation
& Lane-lane and lane-element relations; may ignore geometry quality. \\

\midrule

\multirow{4}{=}{Robustness / Generalization}
& Relative Degradation
& $\Delta_{\mathrm{rel}}=\frac{S_{\mathrm{src}}-S_{\mathrm{tgt}}}{S_{\mathrm{src}}}$
& $S_{\mathrm{src}}$: source-domain score; $S_{\mathrm{tgt}}$: target-domain score
& Adverse-condition evaluation; depends on domain split design. \\
\cmidrule(lr){2-5}

& Rare-class Recall
& $\mathrm{Rec}_{\mathrm{rare}}=\frac{1}{|\mathcal{C}_{r}|}\sum_{c\in\mathcal{C}_{r}}\mathrm{Rec}_c$
& $\mathcal{C}_{r}$: rare-class set; $\mathrm{Rec}_c$: class-wise recall
& Long-tail signs, markings, and signal states; rare-class definitions vary. \\

\midrule

\multirow{4}{=}{Downstream-Oriented}
& Planning / Rule Compliance
& $\mathrm{Comp}=\frac{1}{T}\sum_{t=1}^{T}\mathbb{I}(\pi_t \ \mathrm{satisfies}\ \mathcal{R}_t)$
& $\pi_t$: planned trajectory or action; $\mathcal{R}_t$: applicable road rules or constraints
& Planning and rule-aware driving; hard to isolate perception errors. \\
\cmidrule(lr){2-5}

& Temporal Stability
& $\mathrm{Stab}=1-\frac{1}{T-1}\sum_{t=2}^{T}d(\hat{Y}_t,\hat{Y}_{t-1})$
& $\hat{Y}_t$: prediction at frame $t$; $d(\cdot)$: temporal discrepancy measure
& Video perception and online map updates; may penalize true scene changes. \\

\bottomrule
\end{tabular}%
}
\vspace{-7pt}
\end{table*}

\subsection{Benchmark Fragmentation and Evaluation Gaps}
\label{subsec:fragmentation_gaps}

The dataset and metric landscape above reveals structural gaps in road infrastructure perception. These gaps are not merely limitations of individual benchmarks, but reflect deeper fragmentation in how infrastructure elements are defined, annotated, evaluated, and compared across tasks.

First, task-specific datasets remain fragmented in sensing modality, coordinate system, annotation definition, and benchmark target. Lane datasets often use image-space curves, drivable-area datasets use pixel masks, parking datasets use corners or entrance lines, traffic sign datasets use boxes and class labels, and HD-map datasets use BEV vectors or map elements. Although these designs are reasonable for individual tasks, they make cross-task comparison difficult. For example, performance on 2D lane localization cannot be directly compared with BEV lane-graph prediction, even when both aim to support lane-level planning.

Second, evaluation protocols offer weak cross representation comparability. Metrics with the same name may measure different quantities across output spaces. IoU may refer to image masks, BEV occupancy grids, object boxes, or map-element regions, while F1 may evaluate class recognition, boundary matching, or graph topology. Without representation-aware definitions, benchmark scores can be misleading: high drivable-area mIoU, traffic-light AP, or lane-detection F1 does not necessarily indicate that a system understands where the vehicle can safely and legally move.

Third, existing datasets provide limited infrastructure-level semantics. Most annotations describe visible objects, regions, or local geometry, while fewer encode lane association, valid maneuver constraints, traffic-rule grounding, or route-level semantics. This limitation is especially critical for regulatory-semantic infrastructure. Traffic signs and lights are usually annotated as objects with class or state labels, but benchmarks rarely specify which lane, direction, or maneuver they govern. Road markings are often represented as regions or map polygons, while their legal meanings and interactions with traffic lights, crosswalks, and lane topology remain weakly modeled. As a result, current benchmarks evaluate recognition more than rule-aware understanding.

Fourth, cross-region, long-tail, and adverse-condition evaluation remains insufficient. Road infrastructure varies across countries, cities, weather, maintenance conditions, and regulation systems, including lane marking styles, sign designs, signal layouts, curb structures, and road textures. Rare signs and uncommon markings are particularly important because they may encode safety-critical rules. Although recent datasets improve environmental diversity, most benchmarks still lack unified protocols for evaluating cross-region transfer, long-tail categories, degraded markings, construction zones, adverse weather, and sensor degradation.

Fifth, downstream-oriented evaluation remains underdeveloped. Infrastructure perception ultimately supports localization, online map updating, prediction, planning, and rule-compliant control, yet most benchmarks evaluate isolated perception accuracy rather than behavioral impact. Errors have unequal downstream consequences: a small road-mask error may be harmless, whereas an incorrect lane-traffic-light association or missing stop line can invalidate a maneuver. Future benchmarks should therefore assess whether predicted infrastructure elements improve route-level correctness, map consistency, lane compliance, rule compliance, closed-loop safety, and temporal stability.

A unified benchmark for road infrastructure perception should evaluate geometry, function, semantics, topology, robustness, temporal consistency, and downstream utility. Geometric-structural elements require metric accuracy, continuity, topology, and map compatibility; functional-spatial elements require traversability, uncertainty, parking usability, and planning compatibility; and regulatory-semantic elements require rule grounding, lane relevance, behavior-level consequences, and cross-region semantic generalization. These gaps motivate future directions in unified infrastructure representation, open-world robustness, relation- and rule-aware reasoning, and downstream-oriented evaluation.

% \section{Datasets, Benchmarks, and Evaluation Protocols}
% \subsection{Benchmarking Landscape of Road Infrastructure Perception}
% \subsection{Datasets by Infrastructure Task Family}
% \subsection{Evaluation Protocols and Metrics}
% \subsection{Current Benchmark Gaps and Limitations}

\section{Challenges and Future Directions}
\label{sec:challenges_future}

Despite substantial progress in lane detection, curb and road-boundary extraction, lane topology reasoning, drivable-space perception, parking slot detection, traffic sign recognition, traffic-light state understanding, and road marking recognition, road infrastructure perception remains a fragmented field. Existing studies are often optimized for isolated tasks and task-specific benchmarks, whereas autonomous driving requires infrastructure elements to be jointly interpreted as structured, map-compatible, and behavior-relevant scene components. The key challenge is therefore shifting from improving individual detectors or segmentors toward building unified infrastructure understanding that integrates geometry, traversability, topology, and regulatory semantics.

\subsection{Toward Unified Infrastructure Representation}

Current infrastructure perception is divided across heterogeneous output formats, including image-space masks, point-level labels, BEV grids, vectorized polylines, graph structures, bounding boxes, and semantic-symbolic labels. These representations are effective for individual tasks but difficult to integrate into a common autonomy pipeline. A lane mask, curb point set, drivable-area grid, and traffic-light box may each be accurate under separate metrics, yet remain weakly aligned in a shared spatial and semantic representation.

Future research should therefore move toward unified and interoperable infrastructure representations. A promising direction is a hybrid BEV--vector--graph--symbolic formulation, where BEV features provide metric grounding, vector primitives encode lanes and boundaries, graph structures represent topology and element relations, occupancy fields describe traversability, and symbolic attributes capture traffic rules. Such representations would better connect perception with online mapping, planning, and rule-compliant decision-making.

\subsection{Robustness and Generalization in Open Road}

Road infrastructure is strongly shaped by local regulations, construction standards, road maintenance, and regional driving conventions. Lane markings, curb forms, road textures, traffic signs, signal layouts, parking markings, and rule-bearing road markings may vary substantially across cities and countries. Moreover, many safety-critical infrastructure elements are long-tailed, degraded, occluded, or visible only under difficult illumination conditions. As a result, standard random-split benchmarks often overestimate real-world robustness.

Future datasets and models should explicitly address cross-region, cross-weather, cross-time, and cross-regulation generalization. Evaluation should include worn markings, damaged signs, occluded curbs, temporary construction zones, rare road markings, night scenes, rain, fog, snow, and glare. Model development should also emphasize domain adaptation, uncertainty estimation, temporal aggregation, and map-aware reasoning. For regulatory-semantic elements, open-vocabulary and language-grounded models are promising, but they must be spatially grounded and validated against traffic-rule context rather than treated as generic classifiers.

\subsection{From Element Detection to Rule-Aware Understanding}

Most existing methods still focus on detecting or segmenting individual elements. However, the driving relevance of infrastructure often depends on relationships among elements. A traffic light must be associated with the lane or maneuver it controls; a stop line should be linked to an intersection or signal; a road arrow must be interpreted with lane topology; and a speed-limit sign should be assigned to the corresponding road segment. Independent correct predictions can produce incorrect driving decisions if their relations are not inferred.

Future systems should model road infrastructure as a relation- and rule-aware scene structure. One promising formulation is an infrastructure scene graph, where nodes represent lanes, boundaries, curbs, drivable regions, traffic signs, traffic lights, road markings, parking slots, and intersection components, while edges encode adjacency, connectivity, control relations, regulatory dependencies, and functional constraints. This direction requires joint reasoning over geometric-structural, functional-spatial, and regulatory-semantic infrastructure, enabling lane--signal association, sign--lane grounding, marking--maneuver interpretation, and planning-compatible rule understanding.

\subsection{Downstream-Oriented Evaluation and Benchmarking}

Evaluation remains a major bottleneck. Current metrics are largely task- and representation-specific, such as IoU, F1, AP, boundary distance, topology accuracy, or vector-map mAP. Although useful, these metrics do not always reflect autonomy-level utility. A small geometric error may be harmless in one case but critical if it changes lane connectivity, drivable-space boundaries, traffic-light association, or rule interpretation. Similarly, high segmentation or detection accuracy does not guarantee safe planning, map consistency, or legitimate driving behavior.

Future benchmarks should evaluate infrastructure perception not only by element-level accuracy, but also by downstream utility. Metrics should consider map compatibility, temporal stability, route-level connectivity, planning feasibility, lane compliance, rule compliance, infrastructure change detection, and closed-loop driving performance. A unified benchmark should include multi-sensor observations, BEV annotations, vectorized map elements, lane graphs, drivable-space labels, regulatory elements, temporal sequences, and cross-element relationships. Ultimately, road infrastructure perception should be judged by how reliably its structured outputs support safe, lawful, and map-compatible autonomous driving.

\section{Conclusion}

This survey frames road infrastructure perception as a unified problem for autonomous driving rather than a set of isolated tasks. Centered on geometric organization, functional accessibility, and regulatory semantics, we organize core tasks into geometric-structural, functional-spatial, and regulatory-semantic infrastructure, showing that lane detection, curb and road-boundary detection, lane topology reasoning, drivable-space perception, parking slot detection, traffic sign recognition, traffic-light state understanding, and road marking recognition share a common objective: producing structured road-scene representations for localization, online mapping, planning, and rule-compliant decision-making. Across these tasks, the field is moving from image-space masks, object boxes, and class labels toward BEV grids, vectorized primitives, topology graphs, occupancy fields, and semantic-symbolic representations, because infrastructure outputs must be metrically grounded, topologically consistent, semantically interpretable, and downstream-usable. Existing datasets and metrics remain fragmented across sensing setups, annotation formats, coordinate systems, representations, and task objectives. Future progress requires unified representations that jointly model geometry, traversability, topology, and regulatory semantics, together with evaluation protocols that assess map consistency, planning feasibility, lane compliance, and rule-compliant driving behavior.

{
\bibliographystyle{IEEEtran}
\bibliography{ref}
}

% \newpage

 

% % \bf{If you include a photo:}\vspace{-33pt}
% % \begin{IEEEbiography}[{\includegraphics[width=1in,height=1.25in,clip,keepaspectratio]{fig1}}]{Michael Shell}
% % Use $\backslash${\tt{begin\{IEEEbiography\}}} and then for the 1st argument use $\backslash${\tt{includegraphics}} to declare and link the author photo.
% % Use the author name as the 3rd argument followed by the biography text.
% % \end{IEEEbiography}

% % \vspace{11pt}


\vfill

\end{document}